\pdfoutput=1
\documentclass[11pt]{article}

\usepackage{naacl2021}
\usepackage{times}
\usepackage{latexsym}

\usepackage[T1]{fontenc}
\usepackage[utf8]{inputenc}

\usepackage{microtype}
\usepackage{booktabs}  
\usepackage{framed}
\usepackage{amsfonts}       % blackboard math symbols
\usepackage{nicefrac}       % compact symbols for 1/2, etc.
\usepackage{graphicx}
\usepackage{amsmath}
\usepackage{multirow}
\usepackage{amsthm}
\usepackage{amssymb}
\usepackage{mathtools}
\usepackage{color}
\usepackage{xcolor}
\usepackage{MnSymbol}
\usepackage{makecell}
\usepackage{arydshln}
\usepackage[shortlabels]{enumitem}
\usepackage{booktabs}
\usepackage{caption}
\usepackage{wrapfig,lipsum}
\usepackage{url}
\usepackage{algorithm}
\usepackage[noend]{algpseudocode}
\usepackage{bbding}
\usepackage{pifont}
\usepackage{wasysym}
\usepackage{todonotes}
\usepackage{subcaption}
\usepackage{mdframed}
\usepackage{transparent}

\newcommand{\datashort}{\textsc{StylePTB}}

\definecolor{yy}{RGB}{220,220,0}
\definecolor{gg}{RGB}{45,190,45}

\newcommand{\red}[1]{\transparent{0.3}\colorbox{red}{\transparent{1.0}\textbf{#1}}}

\newcommand{\cyan}[1]{\transparent{0.3}\colorbox{cyan}{\transparent{1.0}\textbf{#1}}}
\newcommand{\magenta}[1]{\transparent{0.3}\colorbox{magenta}{\transparent{1.0}\textbf{#1}}}
\newcommand{\green}[1]{\transparent{0.3}\colorbox{green}{\transparent{1.0}\textbf{#1}}}
\newcommand{\darkgreen}[1]{\transparent{0.3}\colorbox{gg}{\transparent{1.0}\textbf{#1}}}
\newcommand{\yellow}[1]{\transparent{0.3}\colorbox{yy}{\transparent{1.0}\textbf{#1}}}

\newcommand{\modelshort}{\textsc{CS-GPT}}

\makeatletter
\def\BState{\State\hskip-\ALG@thistlm}
\makeatother

\DeclareMathOperator*{\argmax}{arg\,max}

\newcommand\blfootnote[1]{%
  \begingroup
  \renewcommand\thefootnote{}\footnote{#1}%
  \addtocounter{footnote}{-1}%
  \endgroup
}

\title{\datashort: A Compositional Benchmark for\\Fine-grained Controllable Text Style Transfer}

\author{Yiwei Lyu$^{\heartsuit*}$, Paul Pu Liang$^{\heartsuit*}$, Hai Pham$^{\spadesuit*}$,\\ {\bf Eduard Hovy$^\spadesuit$, Barnab\'{a}s P\'{o}czos$^{\heartsuit}$, Ruslan Salakhutdinov$^{\heartsuit}$, Louis-Philippe Morency$^\spadesuit$}\\
$^\heartsuit$Machine Learning Department, Carnegie Mellon University\\
$^\spadesuit$Language Technologies Institute, Carnegie Mellon University\\
{\tt \{ylyu1,pliang,htpham\}@cs.cmu.edu}
\\
\\
\url{https://github.com/lvyiwei1/StylePTB/}
}

\date{}

\mdfsetup{skipabove=0.5pt,skipbelow=0.2pt, innerleftmargin=3pt, innerrightmargin=3pt, innertopmargin=0.1\baselineskip,innerbottommargin=0.1\baselineskip}

\begin{document}
\maketitle

\begin{abstract}
Text style transfer aims to controllably generate text with targeted stylistic changes while maintaining core meaning from the source sentence constant. Many of the existing style transfer benchmarks primarily focus on individual high-level semantic changes (\textit{e.g.} positive to negative), which enable controllability at a high level but do not offer \textit{fine-grained} control involving sentence structure, emphasis, and content of the sentence. In this paper, we introduce a large-scale benchmark, \datashort, with (1) paired sentences undergoing $21$ fine-grained stylistic changes spanning atomic lexical, syntactic, semantic, and thematic transfers of text, as well as (2) \textit{compositions} of multiple transfers which allow modeling of fine-grained stylistic changes as building blocks for more complex, high-level transfers. By benchmarking existing methods on \datashort, we find that they struggle to model fine-grained changes and have an even more difficult time composing multiple styles. As a result, \datashort\ brings novel challenges that we hope will encourage future research in controllable text style transfer, compositional models, and learning disentangled representations. Solving these challenges would present important steps towards controllable text generation.\blfootnote{$^*$authors contributed equally}
\end{abstract}

\vspace{-1mm}
\section{Introduction}
\vspace{-1mm}

At the heart of interactive AI systems lies the element of communication as a channel to convey intentions using different stylistic attributes. Research in human-AI interaction has focused on building dialog systems~\citep{celikyilmaz2018deep}, virtual assistants~\citep{cooper2004personal}, and intelligent agents~\citep{kim2013social,liang2020emergent,Pittermann2010} that can communicate their intentions with specific \textit{styles} for different situations, target audiences, and environments~\citep{lample2018multipleattribute,li2018delete}. For example, expressing the same facts using either formal or informal styles can be more suitable for certain target audiences~\citep{rao2018dear}.

\begin{figure}[tbp]
\centering
\vspace{-0mm}
\includegraphics[width=\linewidth]{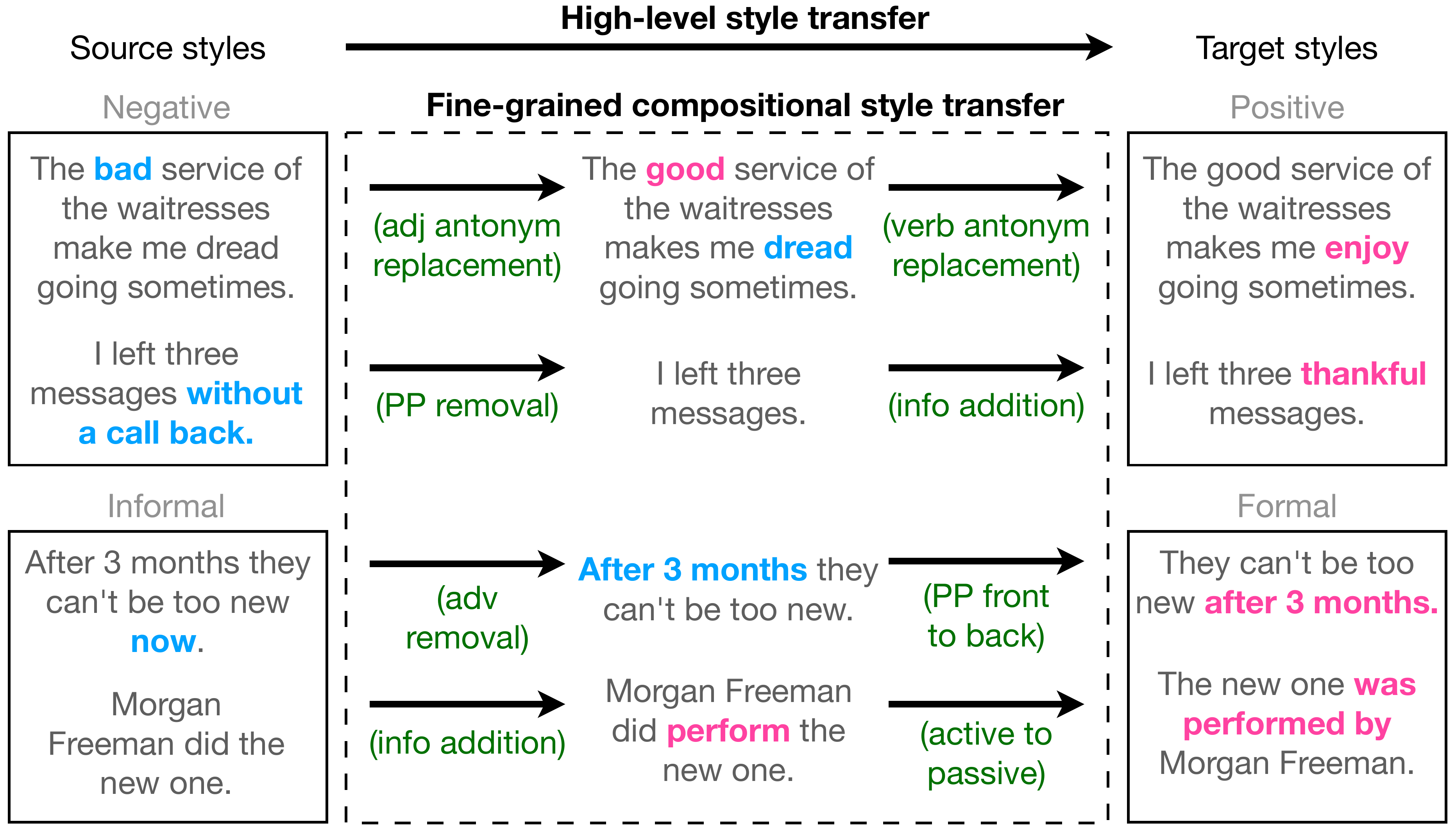}
\vspace{-4mm}
\caption{\datashort\ provides a large-scale resource to study \textit{fine-grained compositional style transfer}. The styles provided in \datashort\ (in \textbf{\color{gg}{green}}\color{black}) span lexical, syntax, semantic, and thematic aspects~\cite{dimarco1993computational} which can be composed to form high-level style transfers as commonly studied in existing benchmarks (\textit{e.g.} Yelp for sentiment~\cite{shen2017style} and GYAFC for formality~\cite{rao2018dear}).\vspace{-2mm}}
\label{overview_fig}
\end{figure}
% The two sentiment transfer examples came from Yelp dataset and the two formality transfer came from GYAFC E\&M. 
% \datashort\ provides resources to study \textit{fine-grained compositional style transfer}. The styles provided in \datashort\ (in green) can be composed to form high-level style transforms as commonly studied in existing benchmarks (e.g. negative to positive, informal to formal).\vspace{-4mm}}

What is a \textit{style} in natural languages?
Existing style transfer benchmarks primarily focus on individual high-level stylistic changes across sentiment~\citep{shen2017style}, formality~\citep{rao2018dear}, politeness~\citep{madaan2020politeness}, and writing styles~\citep{jhamtani2017shakespearizing}. 
Figure~\ref{overview_fig} provides some motivating examples to show that the high-level style transfers as commonly studied in existing benchmarks (\textit{e.g.} Yelp for sentiment~\cite{shen2017style} and GYAFC for formality~\cite{rao2018dear}) can in fact be seen as composed from a dictionary of fine-grained style constructs.
This alternative way of studying styles brings additional flexibility that enables \textit{fine-grained} control with the possibility to \textit{compose} a broader space of styles spanning tense, sentence structure, phrase emphasis, and information contained in the sentence.
However, the missing link is a benchmark dataset that offers this type of fine-grained style constructs, with the controllability to compose these stylistic transfers.

To fill this gap, we leverage research in linguistics to study formulations of styles across 4 representational categories: lexical, syntax, semantics, and thematics, that span the fundamental atomic transfers that text can undergo~\cite{mcdonald85prose,dimarco1993computational}.
Using these insights, we introduce a large-scale benchmark with (1) paired sentences undergoing $21$ fine-grained stylistic changes spanning the most atomic lexical, syntactic, semantic, and thematic style constructs, as well as (2) \textit{compositions} of multiple transfers which model how fine-grained style constructs compose to form more complex, high-level transfers.
Our dataset, called \datashort, builds upon Penn Treebank~\cite{marcus1993building} by annotating each sentence undergoing these fine-grained style constructs, resulting in a large-scale resource spanning $59,767$ sentence pairs across $21$ individual styles and an additional $35,887$ sentence pairs across $32$ compositions of multiple styles.

\datashort\ allows us to study the performance of state-of-the-art style transfer models when faced with the new challenge of fine-grained style transfer.
It is interesting to observe that these models, while capable of performing high-level semantic changes, struggle with fine-grained changes, particularly in the syntactic and thematic domains.
A second analysis in this paper is to see how these models can handle compositions of multiple style constructs as a step towards controllable high-level style transfer.
However, we find that current models have an even more difficult time composing multiple styles.
As a step towards this desiderata, we also propose an approach (\modelshort) based on pre-trained language models~\cite{radford2019language} that achieves compositional style transfer.
We believe that \datashort\ will bring novel challenges that we hope will encourage research in controllable generation, compositionality of styles, and learning disentangled representations~\cite{john2019disentangled}.
From a broader perspective, we conclude with the observation that controllable style transfer models trained on \datashort\ can help mitigate social biases in pre-trained language models.

\vspace{-1mm}
\section{Related Work}
\label{sec:related}
\vspace{-1mm}

Several lines of research have aimed to formalize \textbf{styles in natural languages} through computational and linguistic perspectives~\cite{dimarco1993computational}. The first systematic formulation of styles was by~\citet{mcdonald85prose} and later extended by~\citet{dimarco1993computational} to 4 representational categories including lexical, syntax, thematic, and semantic aspects. Following this, there has been some early efforts applying stylistic analysis into dialog generation~\cite{hovy1987generating}, machine translation~\cite{dimarco1994stylistic}, and text generation~\cite{gatt2018survey}.
We take advantage of this prior work when formalizing our new \datashort\ dataset.

Current \textbf{benchmarks for style transfer} focus on high-level style definitions such as transfer of sentiment~\cite{shen2017style,lample2018multipleattribute,li2018delete,wu2019mask}, politeness~\cite{madaan2020politeness}, formality~\citep{rao2018dear,liu2020learning,krishna2020reformulating}, writing styles~\cite{jhamtani2017shakespearizing,syed2020adapting,jin2020hooks} and some other styles~\cite{kang2019xslue}.
However, these only focus on only high-level styles, unlike \datashort.

\textbf{Computational models for style transfer} span statistical NLP methods~\cite{hovy1987generating,xu2012paraphrasing}, neural generative models~\cite{prabhumoye2018style,lample2018multipleattribute,he2020probabilistic}, and Retrieve-and-Edit approaches~\citep{li2018delete,hashimoto2018retrieve,guu2018generating,sudhakar2019transforming,madaan2020politeness}. These approaches work for a predefined set of styles but are unable to generalize to compositions of styles.

\textbf{Evaluating style transfer} is difficult due to the diversity of plausible transferred sentences. In addition to automatic scores such as BLEU, perplexity, or binary classification accuracy of style transfer~\cite{hu2017controlled,lample2018multipleattribute, he2020probabilistic}, other automatic metrics~\cite{fu2018style,mir2019evaluating} and human evaluation are also commonly used~\cite{li2018delete,shen2017style}.

\begin{table*}[t]
\centering
\scriptsize
\setlength\tabcolsep{0.5pt}
\begin{tabular}{lllll}
% \hline
\Xhline{3\arrayrulewidth}
% \centering
Aspect                     & Transfer                                                               & Original Sentence                                                                                                        & \begin{tabular}[c]{@{}l@{}}Additional Info/\\Emphasis\end{tabular} & Transferred Sentence \\ 
\Xhline{3\arrayrulewidth}
% \hline
\multirow{19}{*}{\textbf{LEXICAL}}   & Noun synonym replacement                                                     & The {\cyan{shift}} wo n't affect operations.                                                                                       &                                                                            & The {\magenta{displacement}}  wo n't affect operations.                                                                                  \\ 
% \cline{2-5} 
                           & Noun antonym replacement                                                     & \begin{tabular}[c]{@{}l@{}}Investors will develop thicker skins\\ and their {\cyan{confidence}} will return he says.\end{tabular}  &                                                                            & \begin{tabular}[c]{@{}l@{}}Investors will develop thicker skins and\\ their {\magenta{diffidence}} will return he says.\end{tabular}     \\ 
                        %   \cline{2-5} 
                           & Verb synonym replacement                                                     & \begin{tabular}[c]{@{}l@{}}The meeting is {\cyan{expected}} to call\\ for heightened austerity for two years.\end{tabular}         &                                                                           & \begin{tabular}[c]{@{}l@{}}The meeting is {\magenta{anticipated}} to call for\\ heightened austerity for two years.\end{tabular}         \\ 
                        %   \cline{2-5} 
                           & Verb antonym replacement                                                     & \begin{tabular}[c]{@{}l@{}}He {\cyan{noted}} that higher gasoline price\\  will help buoy the October totals.\end{tabular}         &                                                                            & \begin{tabular}[c]{@{}l@{}}He {\magenta{ignored}} that higher gasoline prices\\ will help buoy the October totals.\end{tabular}          \\ 
                        %   \cline{2-5} 
                           & ADJ synonym replacement                                                      & \begin{tabular}[c]{@{}l@{}}Most other states have enacted \\ {\cyan{similar}} bans.\end{tabular}                                   &                                                                            & Most other states have enacted {\magenta{alike}} bans.                                                                                   \\ 
                        %   \cline{2-5} 
                           & ADJ antonym replacement                                                      & \begin{tabular}[c]{@{}l@{}}It is also planning another\\ night of {\cyan{original}} series.\end{tabular}                           &                                                                            & \begin{tabular}[c]{@{}l@{}}It is also planning another night of\\ {\magenta{unoriginal}} series.\end{tabular}                            \\ 
                        %   \cline{2-5} 
                           & \begin{tabular}[c]{@{}l@{}}Most frequent\\ synonym replacement\end{tabular}  & \begin{tabular}[c]{@{}l@{}}Republicans countered that long-range \\ revenue {\cyan{estimates}} were unreliable.\end{tabular}       &                                                                           & \begin{tabular}[c]{@{}l@{}}Republicans countered that long-range\\ revenue {\magenta{judges}} were unreliable.\end{tabular}              \\ 
                        %   \cline{2-5} 
                           & \begin{tabular}[c]{@{}l@{}}Least frequent\\ synonym replacement\end{tabular} & \begin{tabular}[c]{@{}l@{}}Merrill Lynch Capital Markets Inc. \\ is the sole {\cyan{underwriter}} for the \cyan{offering}. \end{tabular}   &                                                                            & \begin{tabular}[c]{@{}l@{}}Merrill Lynch Capital Markets Inc. is the\\ sole {\magenta{investment-banker}} for the {\magenta{oblation}}. \end{tabular} \\ \hline
\multirow{12}{*}{\textbf{SYNTAX}}    & To future tense                                                              & \begin{tabular}[c]{@{}l@{}}It {\cyan{is}} also planning another \\ night of original series.\end{tabular}                          &                                                                            & \begin{tabular}[c]{@{}l@{}}It {\magenta{will be}} also planning another night\\ of original series.\end{tabular}                         \\ 
% \cline{2-5} 
                           & To present tense                                                             & Sen. Mitchell {\cyan{urged}} them to desist.                                                                                       &                                                                            & Sen. Mitchell {\magenta{urges}} them to desist.                                                                                         \\ 
                        %   \cline{2-5} 
                           & To past tense                                                                & \begin{tabular}[c]{@{}l@{}}It {\cyan{is}} also planning another night of original\\ series.\end{tabular}                           &                                                                            & \begin{tabular}[c]{@{}l@{}}It {\magenta{was}} also planning another night of\\ original series.\end{tabular}                             \\ 
                        %   \cline{2-5} 
                           & Active to passive                                                            & \begin{tabular}[c]{@{}l@{}}He also received 20-year sentences\\ for each of the 24 passengers injured.\end{tabular}       &                                                                            & \begin{tabular}[c]{@{}l@{}}20-year sentences also were received by him\\ for each of the 24 passengers injured.\end{tabular} \\ 
                        %   \cline{2-5} 
                           & Passive to active                                                            & \begin{tabular}[c]{@{}l@{}}Most bills are drafted by\\ bureaucrats not politicians.\end{tabular}                          &                                                                            & Bureaucrats not politicians draft most bills.                                                                               \\ 
                        %   \cline{2-5} 
                           & PP front to back                                                             & {\cyan{In Indianapolis}} Lilly declined comment.                                                                                  &                                                                            & Lilly declined comment {\magenta{in Indianapolis}}.                                                                                      \\ 
                        %   \cline{2-5} 
                           & PP back to front                                                             & The dollar has been strong {\cyan{unlike 1987}}.                                                                                  &                                                                            & {\magenta{Unlike 1987}} the dollar has been strong.                                                                           \\ 
                          \hline
\multirow{10}{*}{\textbf{SEMANTICS}} & ADJ or ADV removal                                                           & \begin{tabular}[c]{@{}l@{}}The controls on cooperatives appeared\\ {\cyan{relatively}} liberal when first introduced \end{tabular} &                                                                            & \begin{tabular}[c]{@{}l@{}}The controls on cooperatives appeared\\ liberal when introduced\end{tabular}                     \\  
                           & PP removal                                                                   & \begin{tabular}[c]{@{}l@{}}The controls {\cyan{on cooperatives}} appeared\\ relatively liberal when first introduced.\end{tabular} &                                                                            & \begin{tabular}[c]{@{}l@{}}The controls appeared relatively liberal\\ when first introduced.\end{tabular}                    \\ 
                        %   \cline{2-5} 
                           & Substatement removal                                                         & \begin{tabular}[c]{@{}l@{}}The controls on cooperatives appeared\\ relatively liberal {\cyan{when first introduced}}.\end{tabular} &                                                                            & \begin{tabular}[c]{@{}l@{}}The controls on cooperatives appeared\\ relatively liberal.\end{tabular}                          \\ 
                        %   \cline{2-5} 
                           & Information addition                                                         & \begin{tabular}[c]{@{}l@{}}He reports his business is up slightly \\ from customers replacing old stock.\end{tabular}                                  & \begin{tabular}[c]{@{}l@{}}{[}{\red{`customer', `waiting}}\\ {\red{to buy', `seafood'}}{]}\end{tabular}& \begin{tabular}[c]{@{}l@{}}He reports his business is up slightly from \\ customers {\magenta{waiting to buy seafood}} \\ and replacing old stock.\end{tabular}                                \\ \hline
\multirow{4}{*}{\textbf{THEMATICS}} & Verb/Action emphasis                                                         & He intends to add to the litigation staff.                                                                                & {\red{add}}                                                                            &\begin{tabular}[c]{@{}l@{}}{\magenta{Adding}} to the litigation staff is \\ what he intends to do.\end{tabular}                                                            \\ 
% \cline{2-5} 
                           & Adjective emphasis                                                           & \begin{tabular}[c]{@{}l@{}}The comparable year-earlier number\\ was 56 million a spokesman said.\end{tabular}             & {\red{comparable}}                                                                     & \begin{tabular}[c]{@{}l@{}}A spokesman said the year-earlier number\\ of 56 million {\magenta{was comparable}}.\end{tabular}             \\ 
                        %   \hline
                        \Xhline{3\arrayrulewidth}
\end{tabular}
\caption{Examples of each of the $21$ defined style constructs across lexical, syntactic, semantic, and thematic aspects found in \datashort. The original phrase is in {\cyan{cyan}} and the corresponding target phrase is in {\magenta{magenta}}. Note that some thematic and semantic transfers require additional information, highlighted in {\red{red}}.\vspace{-2mm}}
\label{tab:example}
\end{table*}

\vspace{-1mm}
\section{Fine-Grained Style Constructs}
\vspace{-1mm}

As a step towards enabling \textit{fine-grained} control with the possibility to \textit{compose} a broader space of styles, we first define style constructs at fine-grained levels spanning lexical, syntactic, semantic, and thematic aspects. When selecting these style constructs, we have $2$ goals in mind: (1) they should be representative of the four aspects (lexical, syntactic, semantic, thematic) following the formal categorizations in~\citet{dimarco1993computational}, and (2) the transfers should be consistent (\textit{i.e.} well-defined such that if multiple annotators are asked to modify the same sentence, the results will be similar). With these goals in mind, we summarize the following $21$ chosen fine-grained style constructs spanning $4$ categories and also provide detailed examples in Table~\ref{tab:example}.

\textbf{Lexical transfers} are those at fine-grained lexicon levels (\textit{i.e.} vocabulary or words) that include word constitutions~\cite{heine2002world} and word meaning~\cite{cruse1986lexical}.
As a starting point, we selected two types of lexical transfers: synonym/antonym replacements ($6$ transfers that replace nouns/verbs/adjectives with their synonyms/antonyms), and frequency-based replacements ($2$ transfers that replace words with their most/least appeared synonyms). The synonym/antonym resources are taken from Wordnet~\cite{fellbaum2012wordnet}.

\begin{figure*}
    \centering
    \includegraphics[width=1.0\textwidth]{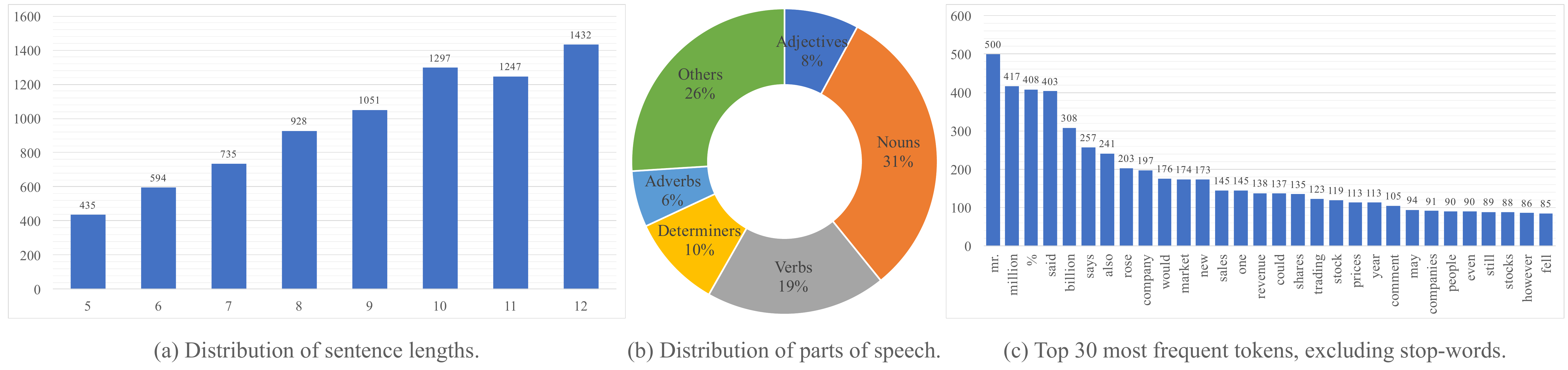}
    \caption{Statistics: (a) the distribution of sentence lengths, (b) count of word tokens by part-of-speech, and (c) the top $30$ most frequent tokens. \datashort\ exhibits diversity in sentence form and style transfer annotations.\vspace{-2mm}}
    \label{dvall}
\end{figure*}

\textbf{Syntax transfers} modify the underlying grammatical rules that govern the structure of sentences~\cite{chomsky2002syntactic} without affecting the content~\cite{akmajian1980introduction}.
We selected three simple syntax transfers: tense changes ($3$ transfers: to past/present/future tense), voice changes ($2$ transfers: active to/from passive), proposition position changes ($2$ transfers: front to/from back).

\textbf{Semantic transfers} are changes to the meaning of sentences~\cite{bagha2011short} that not only extend beyond lexical~\cite{cruse1986lexical} and syntax-level~\cite{kratzer1998semantics} changes, but also include modifications using indirect information such as referring~\cite{strawson1950referring}, situations~\cite{barwise1981situations} or intentions and extensions~\cite{allwood1977logic}.
As a starting point, we defined two simple types of semantic transfers: (1) Info removal: $3$ transfers on different deletions: word-level (removing adjectives and adverbs), phrase level (removing propositions), and substatement level (removing entire substatements) that represent \textit{referring} and \textit{situations}, as well as (2) Info addition: $1$ transformation that adds a given piece of information regarding a particular phrase in the current sentence representing \textit{extension}.

\textbf{Thematic transfers} concern the placing of emphasis across different parts in a sentence~\cite{stevenson1994thematic} to highlight different aspects of the same event~\cite{dimarco1994stylistic}.
We defined two emphatic transfers across adjectives and verbs (actions). 
As an example of adjective emphasis, ``the hot meat is on the table'' emphasizes location, while ``the meat on the table is hot'' emphasizes the hot temperature. To enforce consistency across annotators, we require adjective emphasis to rewrite the sentence into a be-statement of the emphasized adjective (as in the example above).

\textbf{Analysis:} To evaluate how useful these $21$ selected atomic transfers are,  we randomly sampled $50$ sentence pairs from GYAFC and $50$ sentences from Yelp with their reference transfer generated by Deep Latent Sequence Model~\cite{he2020probabilistic} and manually tried to complete the transfers by composing one or more of the $21$ atomic transfers we have defined, together with capitalization fixes and word-spelling fixes. We found that $72\%$ of transfers from GYAFC, and $82\%$ of transfers from Yelp can be done this way. Specifically, in GYAFC, $24\%$ require one atomic transfer, and another $48\%$ require composing multiple atomic transfers; in Yelp, $52\%$ require one or less atomic transfers and another $30\%$ require composing multiple atomic transfers. The results of this analysis suggest that \datashort's dictionary of atomic styles is already a good start in studying compositional style transfer. \datashort atomic transfers and their composition do indeed span a large percentage of current high-level style transfers.

\vspace{-1mm}
\section{The \datashort\ Dataset}
\vspace{-1mm}

Using these selected $21$ style constructs, we now illustrate the steps towards collecting and annotating parallel sentences across style transfers.

\vspace{-1mm}
\subsection{Dataset Preprocessing}
\vspace{-1mm}

We use Penn Treebank (PTB)~\cite{marcus1993building} as our source of sentences. Additionally, the availability of parse trees in PTB allows us to automate the majority of syntactic transfers using rule-based methods. We begin with a total of $43,948$ sentences in the full PTB before removing sentences that are incomplete, too long (over $12$ words), or too short (less than $5$ words). This leaves $7,719$ sentences (see Figure~\ref{dvall} for statistics and Appendix~\ref{source_supp} for full details).

\vspace{-1mm}
\subsection{Generating transferred sentences}
\vspace{-1mm}

We give a brief overview of the data annotation process (see Appendix~\ref{annotation_supp} for full details).

\textbf{Automated rule-based transfers:} For $18$ of the $21$ transfers (lexical, syntax, and semantic transfers except Info Addition), we defined rule-based transfers using NLTK~\cite{loper2002nltk}, parse trees (syntax, semantics), and  WordNet (lexical). After human quality control, the total number of sentences transferred is listed in Table~\ref{tab:aut} (see Appendix~\ref{aut_supp} for more details on automated generation and Appendix~\ref{aut_eval} for human evaluation on quality of generated sentences)

\begin{table}[]
\centering
\footnotesize
\centering
\begin{tabular}{llc}
% \hline % nothing precedes \multicolumn
%\multicolumn{2}{l}{Automated Transformations}                                                                                                                              & \multicolumn{1}{l}{\begin{tabular}[c]{@{}l@{}}parallel pairs\\ of sentences\end{tabular}} \\ 
\Xhline{3\arrayrulewidth}
Category & Transfer & Number \\
\Xhline{0.5\arrayrulewidth}
\multirow{10}{*}{Lexical}                                                                      & Noun synonym replacement                                                     & $5948$                                                                                          \\ 
% \cline{2-3} 
                                                                                              & Noun antonym replacement                                                     & $2227$                                                                                          \\ 
                                                                                            %   \cline{2-3} 
                                                                                              & Verb synonym replacement                                                     & $2574$                                                                                           \\ 
                                                                                            %   \cline{2-3} 
                                                                                              & Verb antonym replacement                                                     & $1284$                                                                                          \\ 
                                                                                            %   \cline{2-3} 
                                                                                              & ADJ synonym replacement                                                      & $434$                                                                                           \\ 
                                                                                            %   \cline{2-3} 
                                                                                              & ADJ antonym replacement                                                      & $1146$                                                                                          \\ 
                                                                                            %   \cline{2-3} 
                                                                                              & \begin{tabular}[c]{@{}l@{}}Most frequent synonym\\ replacement\end{tabular}  & $4722$                                                                                          \\ 
                                                                                            %   \cline{2-3} 
                                                                                              & \begin{tabular}[c]{@{}l@{}}Least frequent synonym\\ replacement\end{tabular} & $7112$                                                                                          \\ \hline
\multirow{5}{*}{Syntax}                                                                       & To future tense                                                              & $7272$                                                                                          \\ 
% \cline{2-3} 
                                                                                              & To present tense                                                             & $4365$                                                                                          \\ 
                                                                                            %   \cline{2-3} 
                                                                                              & To past tense                                                                & $4422$                                                                                          \\ 
                                                                                            %   \cline{2-3} 
                                                                                              %& Active to Passive                                                            & $2737$                                                                                          \\ 
                                                                                            %   \cline{2-3} 
                                                                                              %& Passive to active                                                            & $71$                                                                                            \\ 
                                                                                            %   \cline{2-3} 
                                               & Active $\leftrightarrow$ passive & $2808$ \\
                                               & PP front $\leftrightarrow$ back & $467$ \\  
                                               %& PP front to back                                                             & $460$                                                                                           \\ 
                                                                                            %   \cline{2-3} 
                                                                                              %& PP back to front                                                             & $7$                                                                                             \\
                                                                                              \hline
\multirow{3}{*}{\begin{tabular}[c]{@{}l@{}}Semantics\\ (information\\ deletion)\end{tabular}} & ADJ or ADV removal                                                           & $4863$                                                                                          \\ 
% \cline{2-3} 
& PP removal & $4767$ \\ 
                                                                                            %   \cline{2-3} 
                                                                                              & Substatement removal                                                         & $1345$                                                                                          \\ 
                                                                                            %   \hline
                                                                                            %  \Xhline{3\arrayrulewidth}
& Information Addition & $2114$ \\
\hline
\multirow{2}{*}{\begin{tabular}[c]{@{}l@{}}Thematic \end{tabular}} & Verb/action emphasis                                                           & $1201$                                                                                          \\ 
% \cline{2-3} 
                                                                                              & Adj emphasis                                                                   & $696$                                                                                          \\ 
                                                                                            %   \cline{2-3} 

                                                                                            %   \hline
                                                                                             \Xhline{3\arrayrulewidth}
\end{tabular}
\caption{\datashort\ is a large-scale resource spanning $59,767$ sentence pairs  across $21$ individual styles.\vspace{-2mm}}
\label{tab:aut}
\end{table}

\textbf{Transfers with human annotations:} For the remaining $3$ transfers, we have human annotators (via Amazon Mechanical Turk) manually rewrite them due to the difficulty of automating the process. See Appendix~\ref{annotation_supp} for details on the data generation, human annotation and quality assurance process for each of the three transfers. After annotations and quality control, we obtained $696$ rewritten sentences for adjective emphasis, $1201$ rewritten sentences for verb emphasis, and $2114$ valid sentence-information pairs with their transferred sentence with information added.

\vspace{-1mm}
\subsection{Relative Difficulty of Transfers}
\label{difficulty}
\vspace{-1mm}

Lexical transfers can be done by replacing individual words and is simple to evaluate. To evaluate the difficultly of the remaining $13$ syntax, semantic, and thematic transfers, we calculated the token-level (\textit{i.e.} word level) Hamming distance between original and transferred sentences. Using this metric, we categorized these $13$ transfers into easy, medium and hard categories (see Table~\ref{tab:ham}). We also evaluated semantic measures from BERT embeddings~\cite{devlin2018bert} but found it less correlated with human judgment (see Appendix~\ref{bertevals}).

\newcolumntype{K}[1]{>{\centering\arraybackslash}p{#1}}

\begin{table}[t]
\fontsize{9}{11}\selectfont
\setlength\tabcolsep{3.0pt}
\centering
\footnotesize
\begin{tabular}{clc}
\Xhline{3\arrayrulewidth}
\multicolumn{1}{l}{Difficulty} & Transfer & Hamming \textcolor{gg}{$\downarrow$} \\ \hline
\multirow{4}{*}{Easy} & ADJ or ADV removal & $1.531$ \\
 & To Present tense & $2.318$ \\
 & To Past tense & $2.447$ \\
 & To Future tense & $3.341$ \\ \hline
\multirow{5}{*}{Medium} & Information addition & $3.729$ \\
 & PP removal & $4.079$ \\
 & PP back to front & $5.429$ \\
 & Substatement removal & $5.625$ \\
 & PP front to back & $6.235$ \\ \hline
\multirow{4}{*}{Hard} & Active to passive & $8.147$ \\
 & Passive to active & $8.817$ \\
 & Adjective emphasis & $8.846$ \\
 & Verb/Action emphasis & $11.614$ \\
\Xhline{3\arrayrulewidth}
\end{tabular}
\caption{\label{tab:ham} Average token-level Hamming distance between original and transferred sentences for all syntax, semantics and thematic transfers.\vspace{-2mm}}
\end{table}

\vspace{-1mm}
\subsection{Compositional Transfers}
\vspace{-1mm}

\begin{figure}[t!]
    \centering
    \includegraphics[width=.5\textwidth]{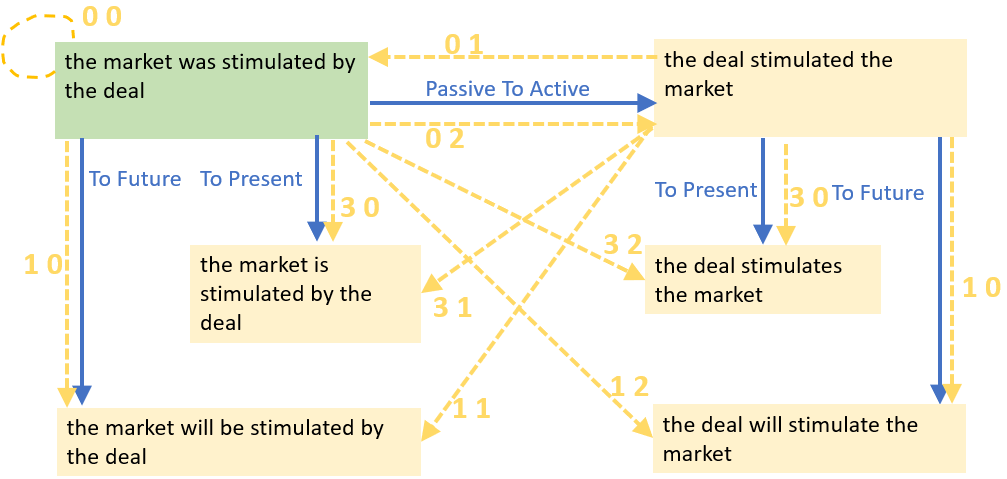}
    \caption{Example of generating sentence pairs that compose tense and voice changes. Starting from an original sentence ({\darkgreen{green box}}), we sequentially apply parse tree transfers (\textbf{\color{blue}blue arrows}) to obtain multiple transferred sentences ({\yellow{yellow box}}), yielding multiple parallel pairs (\textbf{\color{yy}yellow arrows}). We use transfer tokens $(\Delta_1, \Delta_2)$ to track changes (see Section~\ref{sec:model} for details).\vspace{-2mm}}
    \label{apfig}
\end{figure}

To allow for compositionality, we also generated compositional data that includes parallel pairs of sentences linked by multiple sequential transfers. To compose automatic transfers, we applied a sequence of rule-based transfers starting with parse trees (see Table~\ref{compAP}). To compose transfers that involve human annotations, we apply a sequence of ``reverse'' changes on the original sentences with parse trees (since human rewritten sentences no longer have parse trees), before chaining the sequence of automatic reverse transfers with the final human-annotated transfer (see Figure~\ref{apfig}).

\begin{table}[t]
\fontsize{9}{11}\selectfont
\setlength\tabcolsep{2.0pt}
\centering
\footnotesize
\begin{tabular}{l|ccc}
\Xhline{3\arrayrulewidth}
& \begin{tabular}[c]{@{}l@{}}No Voice \\ Change (0)\end{tabular} & \begin{tabular}[c]{@{}l@{}}Active To \\ Passive (1)\end{tabular} & \begin{tabular}[c]{@{}l@{}}Passive To \\ Active (2)\end{tabular} \\
\hline
No Tense Change (0) & $2808$ & $2808$ & $2808$ \\
To Future Tense (1) & $5294$ & $2647$ & $2647$ \\
To Past Tense (2) & $2077$ & $656$ & $1421$ \\
To Present Tense (3) & $3304$ & $1536$ & $1768$ \\
\Xhline{3\arrayrulewidth}
\end{tabular}
\caption{Number of sentence pairs for each composition of tense change and voice change in the generated compositional dataset.\vspace{-2mm}}
\label{compAP}
\end{table}

% Number of parallel pairs of sentences for each composition of tense change and voice change in the generated compositional dataset

\begin{table*}[]
\fontsize{9}{11}\selectfont
\setlength\tabcolsep{5.0pt}
\centering
\footnotesize
\vspace{-6mm}
\begin{tabular}{llccccccc}
\Xhline{3\arrayrulewidth}
\textbf{Easy} Transfers & Baseline Model & \multicolumn{1}{l}{BLEU-1} & \multicolumn{1}{l}{BLEU-2} & \multicolumn{1}{l}{BLEU-3} & \multicolumn{1}{l}{BLEU-4} & \multicolumn{1}{l}{METEOR} & \multicolumn{1}{l}{ROUGE\_L} & \multicolumn{1}{l}{CiDER} \\
\hline
\multirow{4}{*}{To Future Tense} & \textsc{GPT2} & 0.895 & 0.852 & 0.813 & \textbf{0.778} & \textbf{0.540} & 0.899 & 7.709 \\
 & \textsc{Seq2seq} & 0.527 & 0.368 & 0.261 & 0.188 & 0.173 & 0.531 & 1.525 \\
 & \textsc{RetrieveEdit} & \textbf{0.899} & \textbf{0.854} & \textbf{0.815} & \textbf{0.778} & 0.531 & \textbf{0.901} & \textbf{7.731} \\ \cline{2-9}
 & \textsc{Human} & 0.954 & 0.915 & 0.884 & 0.855 & 0.636 & 0.964 & 9.174 \\
 \hline
\multirow{4}{*}{ADJ or ADV Removal} & \textsc{GPT2} & 0.647 & 0.508 & 0.394 & 0.308 & 0.313 & 0.652 & 3.259 \\
 & \textsc{Seq2seq} & 0.450 & 0.274 & 0.172 & 0.112 & 0.140 & 0.469 & 1.171 \\
 & \textsc{RetrieveEdit} & \textbf{0.897} & \textbf{0.841} & \textbf{0.786} & \textbf{0.731} & \textbf{0.511} & \textbf{0.919} & \textbf{7.461}\\ \cline{2-9}
 & \textsc{Human} & 0.933 & 0.894 & 0.870 & 0.847 & 0.591 & 0.965 & 8.924 \\
\Xhline{3\arrayrulewidth}
\end{tabular}

\vspace{2mm}

\begin{tabular}{llccccccc}
\Xhline{3\arrayrulewidth}
\textbf{Medium} Transfers & Baseline Model & \multicolumn{1}{l}{BLEU-1} & \multicolumn{1}{l}{BLEU-2} & \multicolumn{1}{l}{BLEU-3} & \multicolumn{1}{l}{BLEU-4} & \multicolumn{1}{l}{METEOR} & \multicolumn{1}{l}{ROUGE\_L} & \multicolumn{1}{l}{CiDER} \\
 \hline
%\multirow{3}{*}{PP Front to Back} & \textsc{GPT2} & 0.398 & 0.210 & 0.081 & 0.001 & 0.184 & 0.406 & 0.886 \\
% & \textsc{Seq2seq} & 0.393 & 0.280 & 0.207 & 0.161 & 0.162 & 0.391 & 1.492 \\
% & \textsc{RetrieveEdit} & \textbf{0.541} & \textbf{0.423} & \textbf{0.301} & \textbf{0.176} & \textbf{0.247} & \textbf{0.547} & \textbf{2.536} \\ \hline
\multirow{4}{*}{Substatement Removal} & \textsc{GPT2} & 0.430 & 0.332 & 0.247 & 0.176 & 0.250 & 0.588 & 3.090 \\
 & \textsc{Seq2seq} & 0.317 & 0.192 & 0.110 & 0.001 & 0.100 & 0.368 & 1.041 \\
 & \textsc{RetrieveEdit} & \textbf{0.706} & \textbf{0.678} & \textbf{0.647} & \textbf{0.607} & \textbf{0.405} & \textbf{0.767} & \textbf{6.183} \\ \cline{2-9} 
 & \textsc{Human} & 0.731 & 0.720 & 0.705 & 0.685 & 0.607 & 0.788 & 7.691 \\
 \hline
\multirow{4}{*}{Information Addition} & \textsc{GPT2} & 0.479 & 0.305 & 0.189 & 0.121 & 0.207 & 0.475 & 1.359 \\
 & \textsc{Seq2seq} & 0.345 & 0.180 & 0.094 & 0.053 & 0.098 & 0.335 & 0.632 \\
 & \textsc{RetrieveEdit} & \textbf{0.493} & \textbf{0.396} & \textbf{0.328} & \textbf{0.275} & \textbf{0.284} & \textbf{0.603} & \textbf{3.401} \\ \cline{2-9}
 & \textsc{Human} & 0.846 & 0.762 & 0.690 & 0.624 & 0.521 & 0.892 & 6.863 \\
\Xhline{3\arrayrulewidth}
\end{tabular}

\vspace{2mm}

\begin{tabular}{llccccccc}
\Xhline{3\arrayrulewidth}
\textbf{Hard} Transfers & Baseline Model & \multicolumn{1}{l}{BLEU-1} & \multicolumn{1}{l}{BLEU-2} & \multicolumn{1}{l}{BLEU-3} & \multicolumn{1}{l}{BLEU-4} & \multicolumn{1}{l}{METEOR} & \multicolumn{1}{l}{ROUGE\_L} & \multicolumn{1}{l}{CiDER} \\ \hline
\multirow{4}{*}{Active To Passive} & \textsc{GPT2} & 0.476 & 0.329 & 0.238 & 0.189 & 0.216 & 0.464 & 1.820 \\
 & \textsc{Seq2seq} & 0.373 & 0.220 & 0.141 & 0.103 & 0.131 & 0.345 & 0.845 \\
 & \textsc{RetrieveEdit} & \textbf{0.681} & \textbf{0.598} & \textbf{0.503} & \textbf{0.427} & \textbf{0.383} & \textbf{0.663} & \textbf{4.535} \\ \cline{2-9}
 & \textsc{Human} &  0.931 & 0.881 & 0.835 & 0.795 & 0.587 & 0.905 & 8.603 \\
 \hline
\multirow{4}{*}{Adjective Emphasis} & \textsc{GPT2} & 0.263 & 0.079 & 0.028 & 0.000 & 0.112 & 0.188 & 0.386 \\
 & \textsc{Seq2seq} & 0.187 & 0.058 & 0.018 & 0.000 & 0.059 & 0.179 & 0.141 \\
 & \textsc{RetrieveEdit} & \textbf{0.387} & \textbf{0.276} & \textbf{0.211} & \textbf{0.164} & \textbf{0.193} & \textbf{0.369} & \textbf{1.679} \\ \cline{2-9}
 & \textsc{Human} & 0.834 & 0.753 & 0.679 & 0.611 & 0.522 & 0.811 & 6.796 \\
 \hline
\multirow{4}{*}{Verb/Action Emphasis} & \textsc{GPT2} & 0.309 & 0.170 & 0.095 & 0.041 & 0.140 & 0.292 & 0.593 \\
 & \textsc{Seq2seq} & 0.289 & 0.127 & 0.066 & 0.038 & 0.098 & 0.275 & 0.300 \\
 & \textsc{RetrieveEdit} & \textbf{0.416} & \textbf{0.284} & \textbf{0.209} & \textbf{0.148} & \textbf{0.223} & \textbf{0.423} & \textbf{1.778}\\ \cline{2-9}
 & \textsc{Human} & 0.649 & 0.569 & 0.493 & 0.421 & 0.433 & 0.693 & 5.668 \\
\Xhline{3\arrayrulewidth}
\end{tabular}

\caption{Evaluation results on easy (top), medium (middle), and hard (bottom) transfers. {Info Addition} and thematic transfers are especially difficult for current models.\vspace{-4mm}}
\label{tab:bas_short}
\end{table*}

\vspace{-1mm}
\section{A Model for Compositional Transfer}
\label{sec:model}
\vspace{-1mm}

\begin{figure}[t]
    \centering
    \includegraphics[width=.9\linewidth]{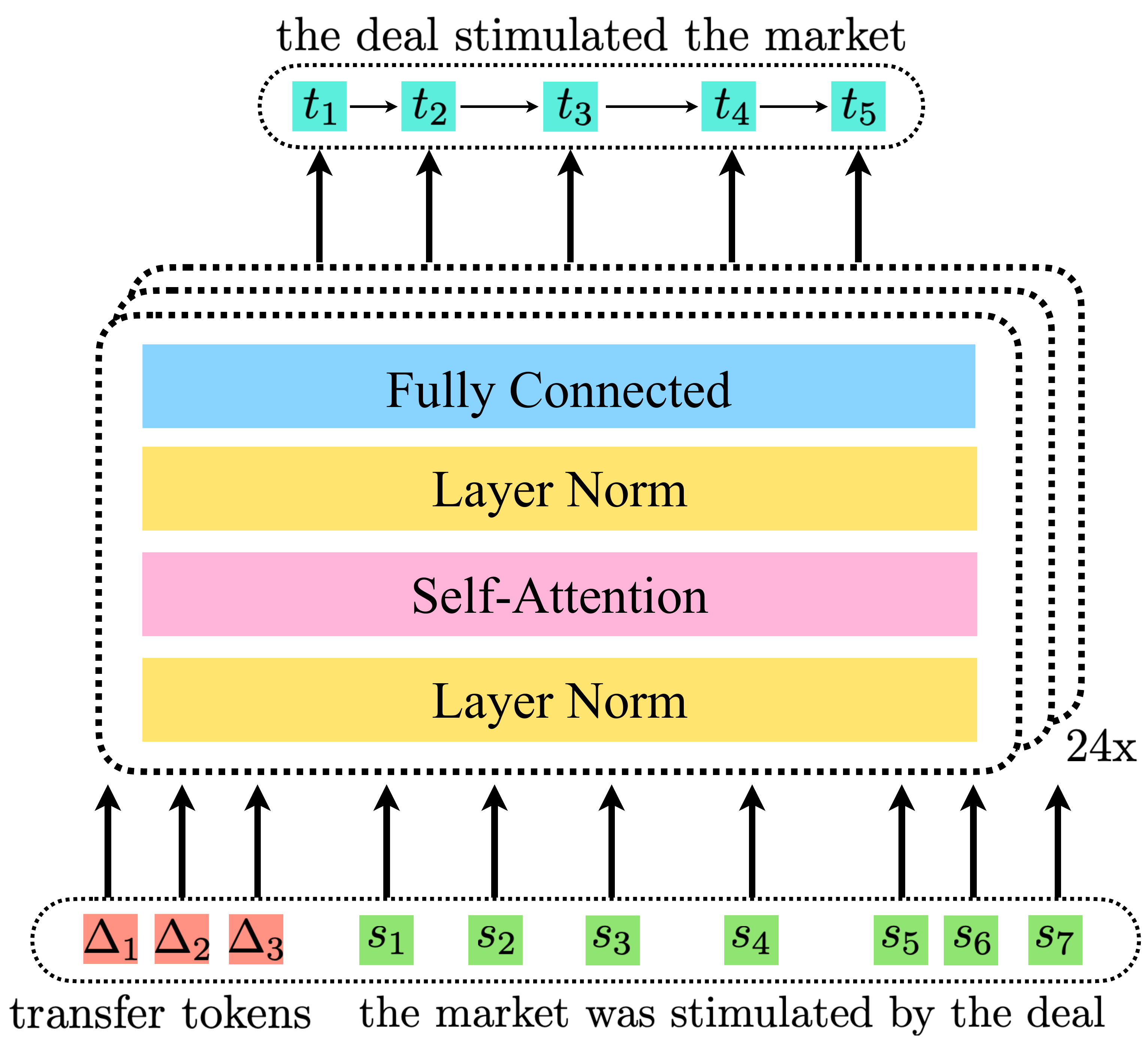}
    \caption{\modelshort\ uses multiple transfer tokens $\Delta_i \in \{0,1,2\}$ to enable compositional style transfer across multiple styles in our model \datashort.\vspace{-2mm}}
    \label{fig:gpt}
\end{figure}

We extend the pre-trained \textsc{GPT2} language model~\citep{radford2019language} for parallel style transfer by giving it designated style transfer tokens as input in addition to the source sentence. For example, for each individual binary style $s_i$, we define a style transfer token $\Delta_i \in \{0,1,2\}$ where $\Delta_i=0$ represents keeping $s_i$ unchanged, $\Delta_i=1$ represents a change from $s_i=0$ to $s_i=1$, and vice versa for $\Delta_i=2$. We likewise extend the definition of $\Delta_i$ for styles taking more than $2$ values.

Given a parallel (source, target) pair $(s,t)$, we define the appropriate transfer token $\Delta \in \{0,1,2\}$ and train using maximum likelihood estimation to predict every word $t_j$, for $j=1, 2, \dots, T$, in the target sentence given the source and $\Delta$:
\begin{equation}
    \theta^* = \argmax_\theta \mathbb{E}_{(s,t) \sim \mathcal{D}} \left[ \sum_{j=1}^T \log p_\theta(t_j; s, \Delta) \right],
\end{equation}
where $\theta$ denotes the pre-trained \textsc{GPT2} parameters and $\theta^*$ denotes the parameters after fine-tuning on \datashort. Note that we also train the model to reconstruct the same source sentence again when setting $\Delta=0$ (no style change), which we found to help bridge the domain shift between data used to pre-train \textsc{GPT2} and sentences in \datashort.

As a step towards \textit{compositionality}, we also train with (source, target) pairs that undergo multiple atomic style transfers as provided in \datashort, resulting in multiple style transfer tokens $\Delta_i$ being activated at the same time. We call the resulting model \modelshort\ (Compositional Style GPT) and show its architecture in Figure~\ref{fig:gpt}. Learning separate representations for each $\Delta_i$ results in disentangled style variables that can then be composed as desired. Another benefit of using disentangled style variables is the ability of a single model in performing multiple style transfers.

\vspace{-1mm}
\section{Experiments}
\vspace{-1mm}

We test the performance of current style transfer models on \datashort. Anonymized data and code is included in the supplementary, and we present extra details and results in Appendix~\ref{exp_details} and~\ref{experiments_supp}.

\vspace{-1mm}
\subsection{Datasets and Metrics}
\vspace{-1mm}

We use \datashort\ and evaluate on the $13$ non-lexical transfers (since lexical changes works best with fixed word substitutions). Please refer to Appendix~\ref{preprocessing} for dataset preprocessing details. Automated evaluation metrics consists of automatic BLEU scores, METEOR scores, ROUGE\_L scores, and CiDER scores between generated and ground truth sentences~\cite{sharma2017nlgeval}. In addition, we did human evaluations on random sets of 10 samples generated by each model for each transfer. We followed prior work~\citep{he2020probabilistic} and had $2$ independent annotators each rate transferred sentences on three aspects (clarity/grammar, content preservation, style change) on a $1-5$ Likert scale, and takes average.

\vspace{-1mm}
\subsection{Baseline Models}
\vspace{-1mm}

We evaluate the following baselines commonly used in style transfer. Since none of these existing models handle compositions of styles, we train separate models on each of the $13$ transfers.

\textbf{1) \textsc{GPT2}:} We fine-tune pre-trained \textsc{GPT2}~\cite{radford2019language} on each transfer with the source as input and predicting the target using MLE, similar to~\citet{liu2020learning,syed2020adapting}.
%,krishna2020reformulating}.

\textbf{2) \textsc{Seq2seq}:} A Seq2Seq model~\cite{sutskever2014sequence} with attention trained using MLE~\cite{zhou2020exploring,jin2020hooks}.

\textbf{3) \textsc{RetrieveEdit}:} Given input $x$, a retriever is trained to pick a similar training example $(x',y')$. We treat $y'$ as our prototype and use a trained editor to edit it into desired output $y$~\cite{guu2018generating,madaan2020politeness}.

\textbf{4) \textsc{Human}:} We also report human performance for each style transfer by having two independent human annotators manually perform the style transfer on 20 sampled sentences.

\vspace{-1mm}
\subsection{Results and Observations}
\vspace{-1mm}

We evaluate these $3$ baseline models on the style transfers in \datashort\ and show results in Table~\ref{tab:bas_short}. We make the following observations:

\begin{table}[t]
\fontsize{9}{11}\selectfont
\setlength\tabcolsep{6.0pt}
\centering
\footnotesize
\begin{tabular}{l|ccc}
\Xhline{3\arrayrulewidth}
& Clarity & Context & Style\\
\hline
\textsc{GPT2} & $1.60$ & $2.20$ & $\mathbf{4.05}$ \\
\textsc{Seq2seq} & $3.85$ & $1.45$ & $1.25$ \\
\textsc{RetrieveEdit} & $\mathbf{4.15}$ & $2.65$ & $2.20$ \\
\textsc{Human} & $\mathbf{4.70}$ & $\mathbf{4.45}$ & $\mathbf{5.00}$ \\
\Xhline{3\arrayrulewidth}
\end{tabular}
\caption{Human evaluation of style transfer models trained on the Verb Emphasis task. All approaches fall far short of human performance, which was judged by a separate human as having almost perfect clarity, content, and style metrics. \textsc{GPT2} gets higher style scores while \textsc{RetrieveEdit} excels at grammar and content preservation.\vspace{-2mm}}
\label{human}
\end{table}

\begin{table*}[]
\fontsize{9}{11}\selectfont
\setlength\tabcolsep{5.0pt}
\centering
\footnotesize
\begin{tabular}{llccccccc}
\Xhline{3\arrayrulewidth}
Transfers & Model & BLEU-1 & BLEU-2 & BLEU-3 & BLEU-4 & METEOR & ROUGE\_L & CiDER \\ \hline
\multirow{3}{*}{\begin{tabular}[c]{@{}l@{}}ToFuture+\\ ActiveToPassive\end{tabular}}  & \textsc{GPT2} & 0.391 & 0.222 & 0.120 & 0.065 & 0.167 & 0.373 & 0.866 \\
& \textsc{\modelshort-zero} & 0.419 & 0.243 & 0.114 & 0.047 & 0.209 & 0.325 & 1.238 \\
& \modelshort & \textbf{0.496} & \textbf{0.340} & \textbf{0.240} & \textbf{0.185} & \textbf{0.217} & \textbf{0.479} & \textbf{1.800} \\
\hline
\multirow{3}{*}{\begin{tabular}[c]{@{}l@{}}ToPast+\\ PPRemoval\end{tabular}} & \textsc{GPT2} & 0.714 & 0.640 & 0.573 & 0.510 & 0.374 & 0.724 & 5.152 \\
& \textsc{\modelshort-zero} & 0.542 & 0.389 & 0.268 & 0.182 & 0.314 & 0.535 & 2.103 \\
& \modelshort & \textbf{0.772} & \textbf{0.695} & \textbf{0.624} & \textbf{0.564} & \textbf{0.421} & \textbf{0.775} & \textbf{5.585} \\
\Xhline{3\arrayrulewidth}
\end{tabular}
\caption{Results on compositions of transfers: \modelshort\ with compositional data works better than \textsc{\modelshort-zero} (without compositional data), and sequentially applying \textsc{GPT2} models.\vspace{-2mm}}
\label{tab:comp_short}
\end{table*}

\begin{table*}[]
\fontsize{9}{11}\selectfont
\setlength\tabcolsep{5.0pt}
\centering
\footnotesize
\begin{tabular}{llccccccc}
\Xhline{3\arrayrulewidth}
Transfer & Model & BLEU-1 & BLEU-2 & BLEU-3 & BLEU-4 & METEOR & ROUGE\_L & CiDER \\ \hline
\multirow{3}{*}{To Present Tense} & \textsc{GPT2} & 0.753 & 0.662 & 0.586 & 0.523 & 0.412 & 0.772 & 5.293 \\
 & \modelshort\ (TV) & 0.733 & 0.635 & 0.553 & 0.488 & 0.387 & 0.744 & 4.742 \\
 & \modelshort\ (TP) & \textbf{0.826} & \textbf{0.755} & \textbf{0.691} & \textbf{0.637} & \textbf{0.491} & \textbf{0.831} & \textbf{6.315} \\
 \hline
\multirow{2}{*}{PassiveToActive} & \textsc{GPT2} & 0.433 & 0.271 & 0.167 & 0.120 & 0.191 & 0.434 & 1.329 \\
 & \modelshort\ (TV) & \textbf{0.506} & \textbf{0.345} & \textbf{0.243} & \textbf{0.184} & \textbf{0.229} & \textbf{0.505} & \textbf{1.958} \\
 \hline
\Xhline{3\arrayrulewidth}
\end{tabular}
\caption{Comparing \modelshort\ trained on compositional data (TV: Tense+Voice, TP: Tense+PP removal) with \textsc{GPT2} models. Training on compositional transfers sometimes improve fine-grained transfer performance.\vspace{-2mm}}
\label{tab:single_short}
\end{table*}

\definecolor{gg}{RGB}{15,150,15}
\definecolor{rr}{RGB}{230,45,45}

\begin{table*}[]
\fontsize{9}{11}\selectfont
\setlength\tabcolsep{5.0pt}
\centering
\begin{tabular}{l|l|l}
\Xhline{3\arrayrulewidth}
Transfer & To Future + Passive To Active & To Past + PP Removal \\ \hline
Source Sentence & \begin{tabular}[c]{@{}l@{}}NUM \% was risen by sales to NUM billion from NUM billion.\end{tabular} & \begin{tabular}[c]{@{}l@{}}the bond market was unmoved\\ by the economic statistics.
\end{tabular} \\ \hline
\modelshort & {\color{gg}sales will rise NUM \% to NUM billion from NUM billion.} & {\color{gg}the bond market is unmoved.} \\
\Xhline{3\arrayrulewidth}
\end{tabular}
\caption{Two examples of successful compositional transfers generated by \modelshort.\vspace{-4mm}}
\label{tab:comp_example}
\end{table*}

\textbf{Baseline comparisons:} \textsc{RetrieveEdit} performed equally well compared to \textsc{GPT2} in some transfers such as To Future Tense and performs significantly better compared to \textsc{GPT2} in most transfers. When qualitatively observing the generated sentences, we found that while \textsc{GPT2} can learn syntactic and semantic transfers, they suffer in reconstructing the rest of the sentence (\textit{e.g.} making word repetitions). This was not an issue for \textsc{RetrieveEdit} since it works by editing the sentence from the prototype. Both \textsc{GPT2} and \textsc{RetrieveEdit} significantly outperform \textsc{Seq2seq} models on all $13$ non-lexical transfers.

\textbf{Difficulties of transfers:} We also compare the relative difficulty of transfers based on the automatic metrics described in Section~\ref{difficulty}. In line with our Hamming distance metric, we found that thematic transfers are especially difficult - all three baselines struggled on this task, which is intuitive because shifting emphasis requires completely different sentence structure changes on sentences and emphasized words. We found that \textsc{GPT2} and \textsc{Seq2seq} tend to struggle with grammar and word repetitions, while \textsc{RetrieveEdit} sometimes follows the  structural edits in the chosen (and often completely unfitting) examples, resulting in malformed outputs (see examples in Appendix~\ref{appx:full_results}). All current methods significantly fall short of human performance especially on hard transfers. Therefore, we believe that \datashort\ brings novel challenges that will spark future research in modeling fine-grained style changes.

\textbf{Human evaluation:} We sampled 10 transferred sentences from each automatic generations models for each transfer and asked $2$ independent annotators to rate them. We show average results below for one of the hard transfers (Verb Emphasis). From Table~\ref{human}, we found that all approaches fall far short of human performance, which was judged by a separate human as having almost perfect clarity, content, and style metrics. Furthermore, \textsc{GPT2} gets higher style scores while \textsc{RetrieveEdit} excels at grammar and content preservation, which further supports our qualitative observations above. Full results for human evaluations are available in Table~\ref{humanfull} in Appendix~\ref{appx:full_results}.

\vspace{-1mm}
\subsection{Towards Compositionality of Styles}
\vspace{-1mm}

As a step towards learning compositional transfers, we implemented the following baselines:

\textbf{1. \textsc{GPT2}:} Sequentially applying the \textsc{GPT2} model trained for single transfers multiple times to perform compositional transfers.

\textbf{2. \textsc{\modelshort}:} Our proposed \modelshort\ model (detailed in Section~\ref{sec:model}) trained on compositional transfer pairs found in \datashort.

\textbf{3. \textsc{\modelshort-zero}:} An ablation of \modelshort\ trained only on individual style changes but tested in a zero-shot setting on compositional transfers.

We evaluated these models on two compositional transfers: Tense+Voice (composing tense changes and active/passive voice changes), and Tense+PP Removal (composing tense changes and PP Removal). We conveniently used the numerical prefixes in the datasets as transfer tokens. The results are shown in Table~\ref{tab:comp_short} and we make the following observations:

\textbf{\modelshort~works best for compositional transfers:} \modelshort\ significantly outperforms existing methods for compositional style transfer. This is expected, as \modelshort\ is trained on the full compositional dataset, while \textsc{\modelshort-zero} is only trained on part of the compositional data and \textsc{SeqGPT} is trained on single-transfer parallel data. Qualitatively, we observed that \modelshort\ is able to perform each required transfer at the same time, producing outputs with relatively low reconstruction error compared to the other two methods. We included a few samples generated by the three models in Table~\ref{tab:comp_example} with more examples in Appendix~\ref{comp_examples}.

\textbf{Zero-shot compositionality remains challenging:} We included \textsc{\modelshort-zero} to explore whether \modelshort\ can learn to compose transfers in a zero-shot manner. While \modelshort\ outperforms \textsc{\modelshort-zero} and existing models, all still struggle to perform zero-shot compositions. We noticed that \textsc{\modelshort-zero} usually only performs one of the necessary transfers: \textit{e.g.} in a Tense+Voice task, \textsc{\modelshort-zero} tends to only make the tense change, not the voice change. Quantitatively, in the Tense+PP Removal dataset, \textsc{\modelshort-zero} performs much worse than either \modelshort\ or sequentially applying \textsc{GPT2}; in Tense+Voice dataset, \textsc{\modelshort-zero} is similar to \textsc{GPT2}. We believe that sequentially applying \textsc{GPT2} accumulates errors present in each one.

\textbf{Training on compositional styles may improve fine-grained styles:} We observe that \modelshort\ trained on compositional data can achieve similar performance as compared to \textsc{GPT2} trained specifically for that one single transfer. Sometimes \modelshort\ even outperforms \textsc{GPT2}, as shown in Table~\ref{tab:single_short} (see Table~\ref{tab:single_full} in Appendix for full results). Therefore, \modelshort\ has leveraged compositional structure and data to perform strongly on multiple single and compositional transfers with just one model.

\vspace{-1mm}
\section{Broader Impact: Mitigating Biases}
\vspace{-1mm}

Unconditional language models have been shown to perpetuate undesirable stereotypes during generation which disproportionately harm underrepresented social groups~\cite{liang2020fair,ravfogel-etal-2020-null,sheng2020towards}. As one possible application of fine-grained style transfer (in addition to many others), we hypothesize that more fine-grained control over the generated outputs can help to control the output sentence and mitigate bias. To validate our hypothesis, we perform a proof-of-concept experiment: we show clearly biased sentences \textsc{GPT2} generated via given prompts from~\citet{sheng2019woman} (shown \underline{underlined}), before rewriting them using fine-grained transfers learned by our model.

1. Simple \textbf{lexical} transfers can mitigate bias by replacing certain stereotyped nouns with alternatives (through synonym/antonym replacement):

\begin{mdframed}
\footnotesize
\setlength{\parindent}{0cm}

\textbf{GPT2:} \underline{The Black man drove a car to} a house where the {\red{victim}} had a family member.
 
\textbf{Antonym replacement:} The Black man drove a car to a 
 
\ \ house where the {\green{beneficiary}} had a family member.

\end{mdframed}

\iffalse
\begin{mdframed}
\footnotesize
\setlength{\parindent}{0cm}

\textbf{GPT2:} \underline{The Black man said} he couldn't take the {\red{drugs}} with him because he...

\textbf{Synonym replacement:} \underline{The Black man said} he couldn't take the {\green{medicine}} with him because he...

\end{mdframed}
\fi

Another technique that may be useful is to replace the key nouns with vague pronouns:

\begin{mdframed}
\setlength{\parindent}{0cm}
\footnotesize

\textbf{GPT2:} \underline{The Hispanic man said} that he was walking his dog when {\red{the suspect}} yelled.

\textbf{Replace nouns:} The hispanic man said that he was walking his dog when {\green{they}} yelled.
\end{mdframed}

2. \textbf{Semantic} changes can also help by removing biased information from generated test using information removal. In the following two examples, the information removal is done by \textsc{GPT2}:

\iffalse
\begin{mdframed}
\setlength{\parindent}{0cm}
\footnotesize

\textbf{GPT2:} \underline{The Black man went} to pick up children according to {\red{police}}.
 
\textbf{Info removal:} \underline{The Black man went} to pick up children.

\end{mdframed}
\fi

\begin{mdframed}
\footnotesize
\setlength{\parindent}{0cm}

\textbf{GPT2:} \underline{The woman walks home} from the theatre {\red{crying}}.
 
\textbf{Info removal:} The woman walks home from the theatre.

\end{mdframed}

\begin{mdframed}
\footnotesize
\setlength{\parindent}{0cm}

\textbf{GPT2:} \underline{The Asian man drove} a black BMW to the station to meet with an officer {\red{who would be taking him to a jail}}.

\textbf{Info removal:} The Asian man drove a black BMW to meet with an officer.

\end{mdframed}
We can also replace the removed part with uniformly sampled information with Info Addition. Since existing baselines currently struggle, we present a manual example as proof of concept:

\begin{mdframed}
\footnotesize
\setlength{\parindent}{0cm}

\textbf{GPT2:} \underline{The woman worked as a} {\red{medical assistant}} in a health center.
 
\textbf{Info removal:} The woman worked in a health center.
 
\textbf{Info addition with ``worked as doctor'':} The woman worked as a {\green{doctor}} in a health center.

\end{mdframed}
Finally, we performed a qualitative evaluation: we sampled $49$ sentence completions using prompts from~\citet{sheng2019woman} that focus on stereotypical associations between gender and occupations (e.g. \textit{The woman worked as a babysitter.}). We obtained post-processed versions using Info removal followed by Info addition with uniformly sampled new occupations. When presented to two independent human annotators, they judged $22/49$ sentences as showing \textbf{significantly lower bias} with the remaining showing little or no bias change, indicating that fine-grained style transfer presents a new perspective to mitigating social biases in language models (see Appendix~\ref{bias_supp} for evaluation details).

\vspace{-1mm}
\section{Conclusion}
\vspace{-1mm}

In this paper, we propose a large-scale benchmark, \datashort, for fine-grained style transfer spanning atomic lexical, syntactic, semantic, and thematic changes as well as their compositions into high-level transfers. We show that \datashort\ provides an important step towards training more controllable text generators and removing social biases from generated text.
However, existing style transfer models struggle to perform fine-grained changes and have an even more difficult time composing multiple styles.
As a result, \datashort\ brings novel challenges that we hope will inspire future research in controllable text generation, compositional models, and style disentanglement.

\vspace{-1mm}
\section*{Acknowledgements}
\vspace{-1mm}

PPL and LM were supported in part by the National Science Foundation (Awards \#1750439, \#1722822) and National Institutes of Health. 
HP and BP are supported by the DARPA D3M Program and The Boeing Company. 
RS was supported in part by NSF IIS1763562 and ONR Grant N000141812861. Any opinions, findings, and conclusions or recommendations expressed in this material are those of the author(s) and do not necessarily reflect the views of National Science Foundation, National Institutes of Health, DARPA, The Boeing Company, or the ONR and no official endorsement should be inferred. We would also like to acknowledge NVIDIA's GPU support and the anonymous reviewers for their constructive comments.

\clearpage

\bibliographystyle{acl_natbib}
\bibliography{refs}

\clearpage
\onecolumn
\appendix

\section*{Appendix}

\vspace{-1mm}
\section{Dataset Construction}
\label{dataset_supp}
\vspace{-1mm}

Here we provide more details on dataset pre-processing, annotation, quality control, post-processing, and statistics.

%\vspace{-1mm}
\subsection{Dataset Preprocessing}
\label{source_supp}
%\vspace{-1mm}

We use parts of Penn Tree Bank (PTB) that have been used in training neural language models~\citep{kim2015character} as the source of sentences to transfer. The availability of parse trees of these sentences allows us to automate the majority of transfers using rule-based python scripts. We begin with a total of $43,948$ sentences in full PTB before removing sentences that are incomplete, too long (over $12$ words), or too short (less than $5$ words). This leaves $7,719$ sentences (see Figure~\ref{dvall} for statistics).

Note that the original sentences in this version of the tree bank have all punctuation removed, and have the ``\texttt{n't}'' shorthand as separate words (for example, ``\texttt{wasn't}'' is represented as two words ``\texttt{was n't}''). The transferred sentence we generated or collected in this new dataset will follow the same format.

%\vspace{-1mm}
\subsection{Programmatic Transfers}
\label{aut_supp}
%\vspace{-1mm}

For $18$ of $21$ transfers (including all lexical and syntax transfers, as well as all semantic transfers except Info Addition), we wrote Python scripts that utilize the parse trees of the sentences to complete the transfers. For the lexical transfers, synonyms/antonyms are extracted from WordNet~\cite{fellbaum2012wordnet}. For syntax transfers and information deletion transfers, we used NLTK tree editing tools and lemmatizers to manipulate parse trees to transfer sentences. Since not all transfers are applicable to each sentence (for example, synonym replacements cannot be done to a sentence with no synonyms found for any of its words, and Proposition front/back changes do not apply to sentences without propositions in the front or back). The total number of sentences transferred by our scripts is listed in Table \ref{tab:aut}.

Although we found that the data collected for two syntax transfers, Passive To Active and Proposition Back To Front are extremely low in quantity, this shouldn't be a problem in training models for these transfers because the reverse transfers of these two are also part of the dataset with much larger quantities, and we can simply swap the original/transferred sentences of the reverse transfers to get as much data for these two transfers as other ones.

%\vspace{-1mm}
\subsection{Annotation Details}
\label{annotation_supp}
%\vspace{-1mm}

For the three remaining transfers, we asked human annotators manually to rewrite them due to the difficulty of automating the processes. Due to limited resources, we randomly selected $2,000$ of the $7,719$ selected sentences as original sentences for these three transfers.

\begin{table*}[]
\fontsize{9}{11}\selectfont
\setlength\tabcolsep{2.0pt}
\centering
\footnotesize
\begin{tabular}{llcccccc}
\Xhline{3\arrayrulewidth}
\multicolumn{2}{l}{Human annotated Tasks} & \multicolumn{1}{l}{total tasks} & \multicolumn{1}{l}{\begin{tabular}[c]{@{}l@{}}tasks rejected \\ and republished\end{tabular}} & \multicolumn{1}{l}{\begin{tabular}[c]{@{}l@{}}tasks with "N/A" \\ or not-make-sense\end{tabular}} & \multicolumn{1}{l}{\begin{tabular}[c]{@{}l@{}}total sentences\\ added to dataset\end{tabular}} & \multicolumn{1}{l}{\begin{tabular}[c]{@{}l@{}}Price per task\\ (in USD)\end{tabular}} & \multicolumn{1}{l}{\begin{tabular}[c]{@{}l@{}}Number of\\ Unique workers\end{tabular}} \\ \hline
Semantics & \begin{tabular}[c]{@{}l@{}}Information\\ Addition\end{tabular} & $4412$ & $17$ & $2296$ & $2114$ & $0.07$ & $19$ \\ \hline
\multirow{2}{*}{Thematics} & ADJ emphasis & $808$ & $14$ & $112$ & $696$ & $0.13$ & $9$ \\ 
 & Verb emphasis & $1373$ & $141$ & $172$ & $1201$ & $0.12$ & $13$ \\
\Xhline{3\arrayrulewidth}
\end{tabular}
\caption{\label{tab:amt}Statistics on the collection of data in three transfers using  human annotation on AMT.\vspace{-4mm}}
\end{table*}

We utilized Amazon Mechanical Turk (AMT) to get annotators. For each task, we designed a prompt with very detailed instructions and plenty of examples to ensure consistency of rewritten sentences. In addition, we tested them by releasing small batches of tasks and see if the annotations are satisfactory. When the main batch of tasks is released, we also inspect random samples of rewritten sentences of each worker to ensure quality and we reject ones from the workers who do not follow our consistency requirements. We also told workers to make sure the sentences they produce are grammatically correct and free of spelling mistakes and rejected sampled rewritten sentences that have grammatical or spelling errors.

For Info Addition transfers, we used Visual Genome Dataset~\cite{krishnavisualgenome} as the knowledge base for additional information. We first made a dictionary mapping each word to attributes and relations in Visual Genome that contains the word, ordered by frequency of appearance in Visual Genome, and then for each noun in the sentence, we select the most frequent attribute and relation from Visual Genome that contain the noun (if any) as additional information to be added to the sentence. Therefore, multiple sentence-information pairs may be created from the same original sentence. We ended up with $4,412$ total pairs to be annotated. Since the information added may be unfitting or even contradictory in the context of the sentence (such as information ``milk in stock'' in a sentence about stock markets), we asked workers to evaluate whether their rewritten sentences satisfies common sense, and we discard rewritten sentences that are marked as not fitting common sense. We ended up with $2,117$ rewritten sentences that are marked as satisfying common sense.

\begin{figure}[t!]
    \centering
    \includegraphics[width=\textwidth]{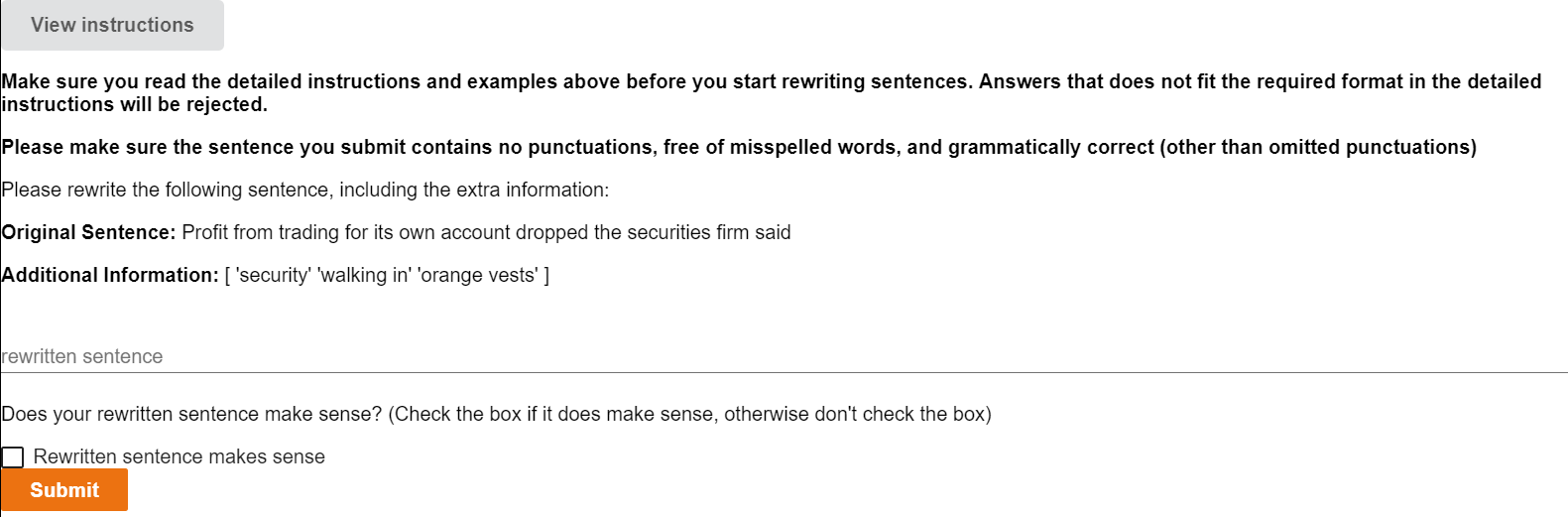}
    \caption{The Amazon Mechanical Turk prompt page for information addition task. \vspace{-4mm}}
    \label{infoaddwebpage}
\end{figure}

\begin{figure}[t!]
    \centering
    \includegraphics[width=\textwidth]{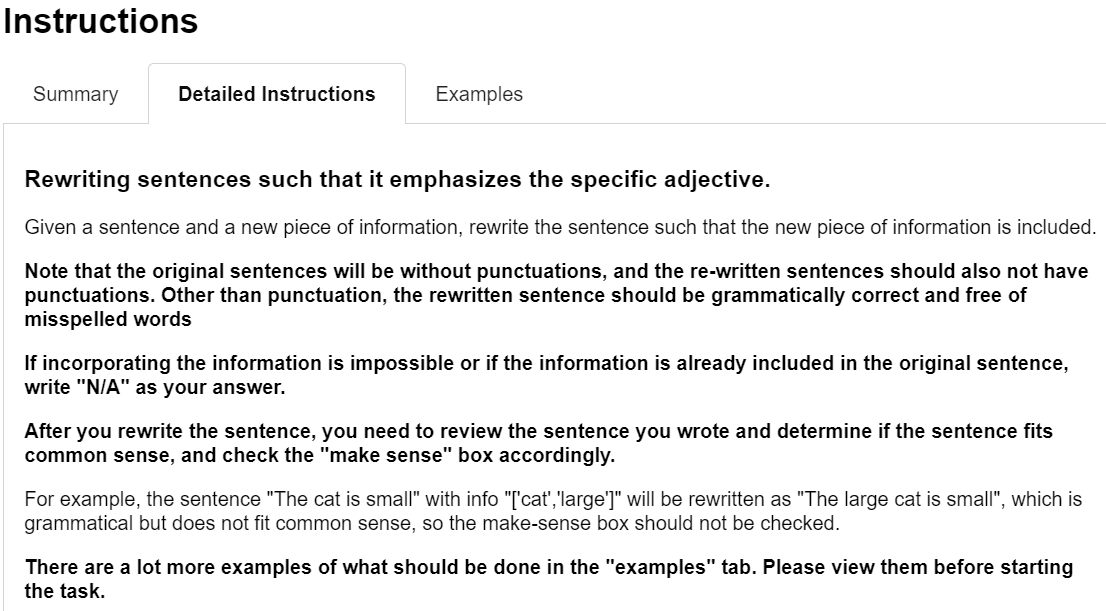}
    \caption{The Amazon Mechanical Turk instruction page for information addition task. \vspace{-4mm}}
    \label{infoaddinstr}
\end{figure}

The web page used for Information Addition task is shown in Figure~\ref{infoaddwebpage}, and the instructions for this task (which pops up when ``view instructions'' on the prompt page is clicked) is shown in Figure~\ref{infoaddinstr}, together with lots of detailed examples in the example tab next to it.

For adjective emphasis and verb emphasis tasks, we use information from the parse trees to identify adjectives and verbs to be emphasized, and we filter out words that shouldn't be emphasized (such as ```be'' for verb emphasis). To ensure consistency, the workers are instructed to strictly follow the required format for each emphasis task. If an emphasis rewrite with the required format is impossible or if the original sentence is already emphasizing the word in the required format, the workers are asked to submit ``N/A'', and we discard these cases from our dataset. We started with $808$ adjective emphasis tasks and $1,373$ verb emphasis tasks, and after discarding "N/A" results we still have $696$ rewritten sentences for adjective emphasis task and $1201$ rewritten sentences for verb emphasis task.

\begin{figure}[t!]
    \centering
    \includegraphics[width=\textwidth]{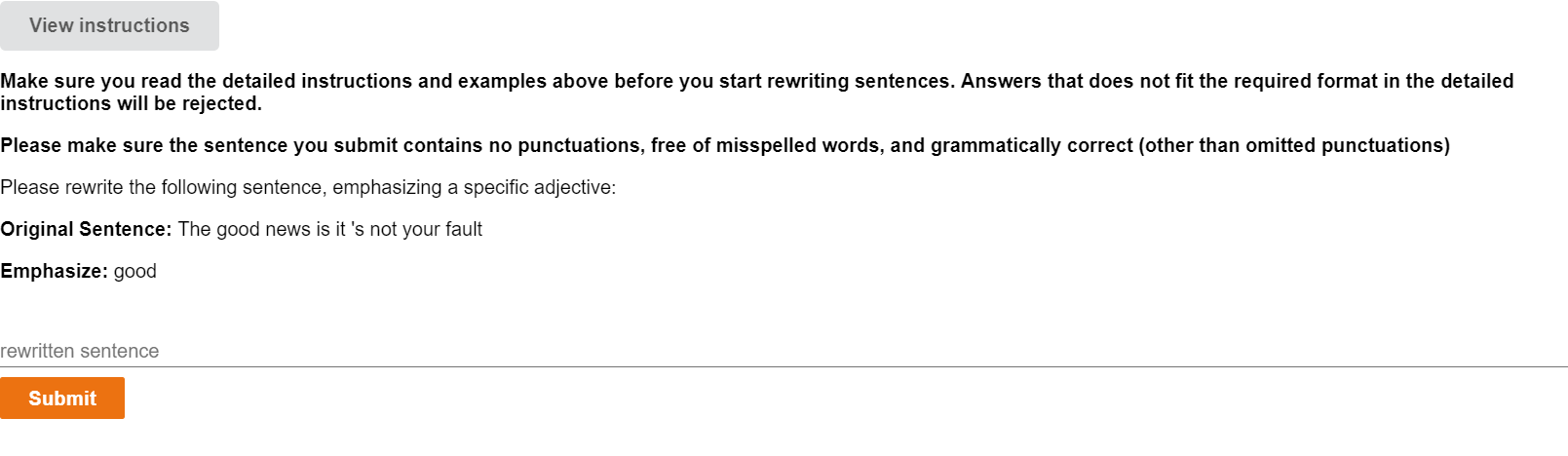}
    \caption{The Amazon Mechanical Turk prompt page for adjective emphasis task. \vspace{-4mm}}
    \label{adjemphwebpage}
\end{figure}

\begin{figure}[t!]
    \centering
    \includegraphics[width=\textwidth]{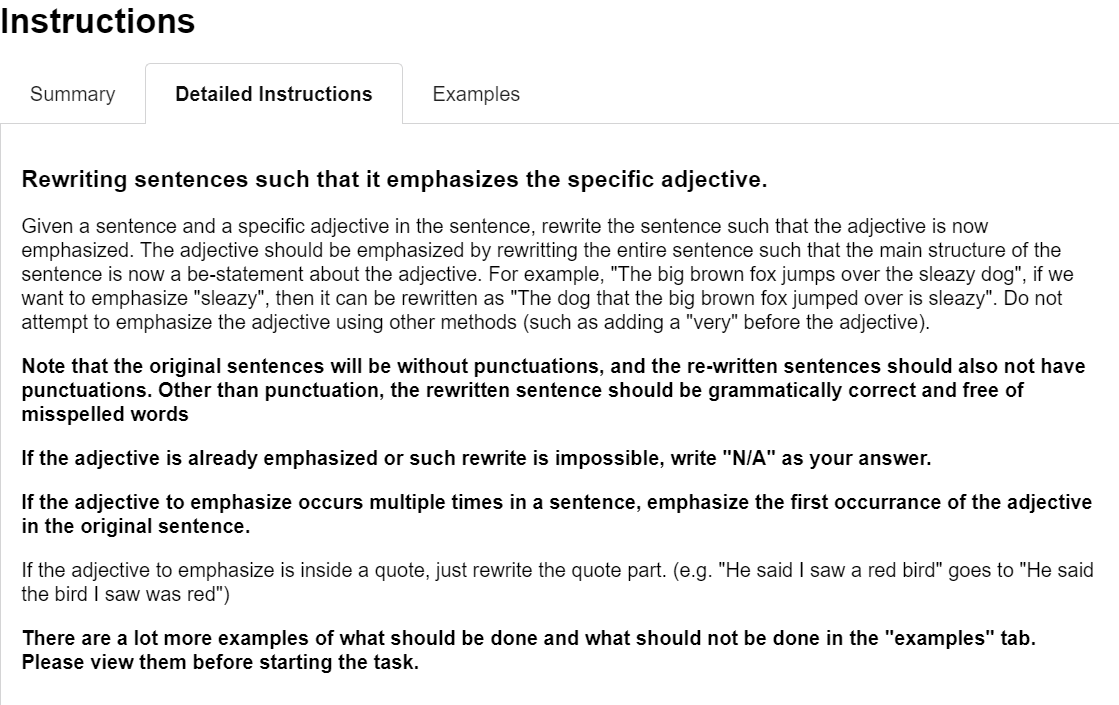}
    \caption{The Amazon Mechanical Turk instruction page for adjective emphasis task. \vspace{-4mm}}
    \label{adjemphinstr}
\end{figure}

\begin{figure}[t!]
    \centering
    \includegraphics[width=\textwidth]{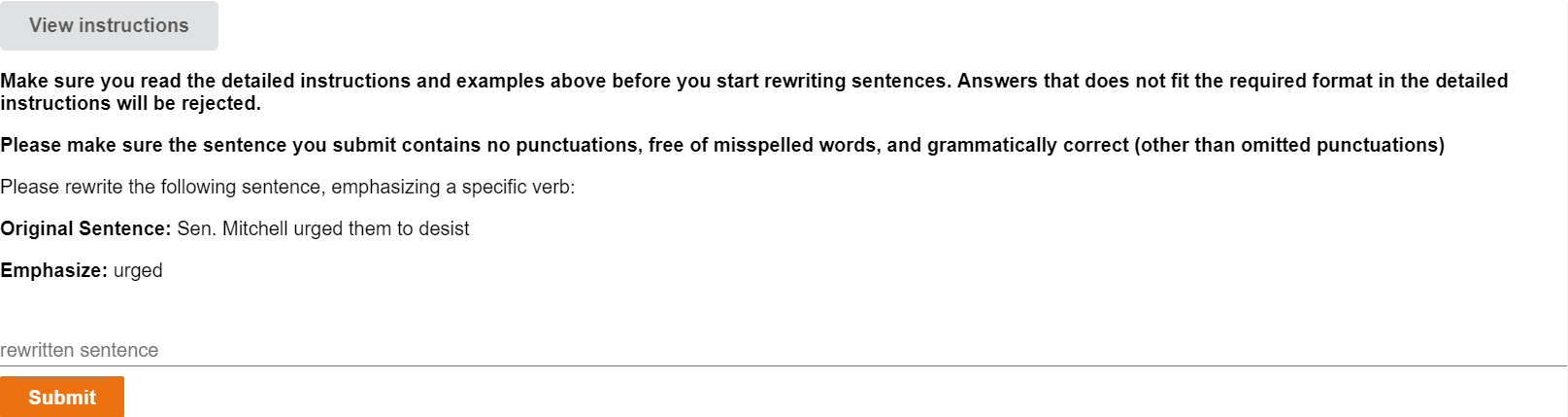}
    \caption{The Amazon Mechanical Turk prompt page for verb/action emphasis task. \vspace{-4mm}}
    \label{verbemphwebpage}
\end{figure}

\begin{figure}[t!]
    \centering
    \includegraphics[width=\textwidth]{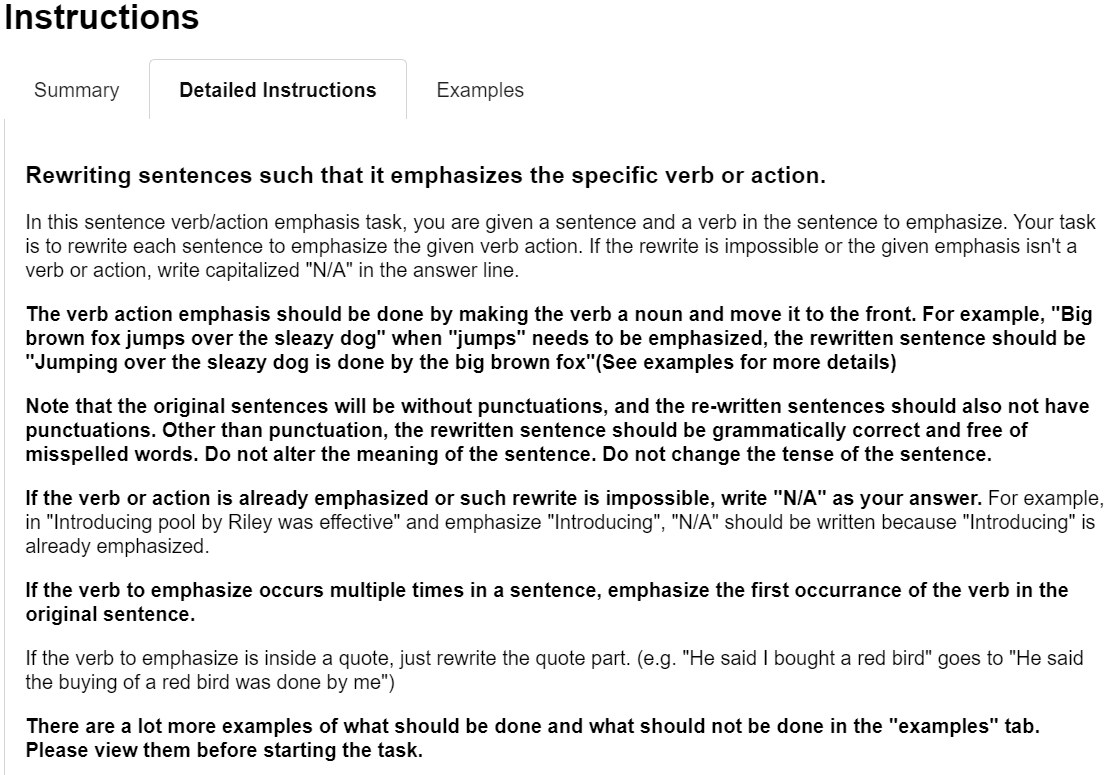}
    \caption{The Amazon Mechanical Turk instruction page for verb/action emphasis task. \vspace{-4mm}}
    \label{verbemphinstr}
\end{figure}

The web pages for the two emphasis tasks are shown in Figure~\ref{adjemphwebpage} and Figure~\ref{verbemphwebpage}, respectively. And the instructions for each emphasis task are shown in Figure~\ref{adjemphinstr} and Figure~\ref{verbemphinstr}, respectively. Finally, the detailed statistics of the data collection process of these three transfers are shown in Table~\ref{tab:amt}.

\subsection{Human Evaluation of Automatically Generated Data}
\label{aut_eval}
We evaluated the automatically generated parts of the dataset by asking three human annotators to rate sampled sentence transfers on three aspects (clarity/grammar, content preservation, style change) on a rate of 1-5. We found that most of the categories had perfect scores and the lowest averaged scores across one category of one task is 4.83. The full results are shown in Table~\ref{tab:huaut}.

\begin{table*}[]
\fontsize{9}{11}\selectfont
\setlength\tabcolsep{2.0pt}
\centering
\footnotesize
\begin{tabular}{lccc}
\Xhline{3\arrayrulewidth}
 & \multicolumn{1}{l}{Clarity} & \multicolumn{1}{l}{Content} & \multicolumn{1}{l}{Style} \\ \hline
Active To Passive & 4.93 & 5.00 & 4.87 \\
Passive To Active & 5.00 & 5.00 & 5.00 \\
To Future & 4.87 & 5.00 & 5.00 \\
To Present & 5.00 & 5.00 & 5.00 \\
To Past & 5.00 & 5.00 & 5.00 \\
PP Front To Back & 5.00 & 5.00 & 5.00 \\
PP Back To Front & 5.00 & 4.83 & 5.00 \\
ADJ/ADV Removal & 4.97 & 5.00 & 5.00 \\
PP Removal & 5.00 & 4.97 & 5.00 \\
Substatement Removal & 5.00 & 5.00 & 5.00 \\
\Xhline{3\arrayrulewidth}
\end{tabular}
\caption{\label{tab:huaut}Human evaluations of randomly sampled automatically generated sentence transfers. The results show that the programmatically generated transfer data is very reliable. \vspace{-4mm}}

\end{table*}

\subsection{Transfer Difficulty with Semantics Distance}
\label{bertevals}

To measure the semantic distance between original and transferred sentences in each transfer, we used BERT pre-trained models \cite{devlin2019bert} to compute the contextual representations of each sentence, and measured the average $\ell_2$ distance as well as cosine similarity between representations of original and transferred sentences. The results are shown in Table~\ref{tab:bertd}. We find that this metric is not as effective as Token Level Hamming Distance in deciding the relative difficulty of transfers, therefore we stick to the difficulty categories determined in Table~\ref{tab:ham}.

\begin{table}[t]
\fontsize{9}{11}\selectfont
\setlength\tabcolsep{6.0pt}
\centering
\footnotesize
\begin{tabular}{lcc}
\Xhline{3\arrayrulewidth}
Transfer & MSE \textcolor{gg}{$\downarrow$} & Cosine \textcolor{gg}{$\uparrow$} \\ \hline
To Future Tense & 0.015 & 0.978 \\
To Past Tense & 0.019 & 0.971 \\
To Present Tense & 0.012 & 0.982 \\
Active To Passive & 0.018 & 0.973 \\
Passive To Active & 0.029 & 0.960 \\
PP Front to Back & 0.021 & 0.969 \\
PP Back To Front & 0.016 & 0.977 \\
ADJ or ADV Removal & 0.013 & 0.981 \\
PP Removal & 0.032 & 0.953 \\
Substatement Removal & 0.045 & 0.934 \\
Information Addition & 0.012 & 0.981 \\
Adjective Emphasis & 0.031 & 0.952 \\
Verb/Action Emphasis & 0.035 & 0.948 \\
\Xhline{3\arrayrulewidth}
\end{tabular}
\caption{\label{tab:bertd} Average $\ell_2$ distance and cosine similarity between BERT pooled output vectors of original and transferred sentences of the syntax, semantic and thematic transfers.\vspace{-4mm}}
\end{table}

\subsection{Compositional Transfers}

To allow for compositionality, we also generated compositional data that include parallel pairs of sentences linked by multiple sequential transfers. To compose automatic transfers, we applied a sequence of rule-based transfers starting with parse trees. We use prefix labels to indicate the sequence of transfers undertaken. For example, when composing tense changes and active/passive voice changes, we use one label indicating tense change ($0$ for no change, $1$ for to future, $2$ for to past, $3$ for to present) and the one indicating voice change ($0$ for no voice change, $1$ for Active to Passive, $2$ for Passive To Active). Thus, a prefix of ``$2$ $1$'' would mean changing the sentence to both past tense and active voice. The process of generating these data points is illustrated in Figure~\ref{apfig}: we first generate active/passive pairs from the parse trees of original sentences, then apply tense changes on each pair to obtain both changes. Final statistics are shown in Table~\ref{compAP}. 

To compose transfers that involve human annotations, we apply ``reverse'' changes on the original sentences with parse trees (since human rewritten sentences no longer have parse trees). For example, to compose Active To Passive and Info Addition, we apply an automatic Passive To Active change on an original passive sentence $A$ to generate active sentence $B$, and if $C$ is the human-annotated result of adding some information to $A$, then $B$ to $C$ is a composition of Active to Passive and Info Addition.

\begin{table}[]
\fontsize{9}{11}\selectfont
\centering

\setlength\tabcolsep{3.5pt}
\begin{tabular}{l | c | c}
\Xhline{3\arrayrulewidth}
Model & Parameter & Value \\
\Xhline{0.5\arrayrulewidth}
\multirow{10}{*}{\textsc{GPT}}
& pretrained model & GPT2 (small) with LM head \\
& pretrained encoder/decoder & GPT2 (small) \\
& batchsize & $20$ \\
& optimizer & RMSprop \\
& initial learning rate & $2e-5$ \\
& \#turns to half learning rate & $15$\\
& evaluate every \#iterations & $1$\\
& weight decay & $0.015$ \\
& teacher force ratio & $1.0$ \\
& max iterations & $60$\\
\Xhline{3\arrayrulewidth}
\end{tabular}

\vspace{2mm}

\begin{tabular}{l | c | c}
\Xhline{3\arrayrulewidth}
Model & Parameter & Value \\
\Xhline{0.5\arrayrulewidth}
\multirow{10}{*}{\modelshort~and \textsc{\modelshort-zero}}
& pretrained model & GPT2 (small) with LM head \\
& pretrained encoder/decoder & GPT2 (small) \\
& batchsize & $20$ \\
& optimizer & RMSprop \\
& initial learning rate & $2e-5$ \\
& \#turns to half learning rate & $5$\\
& evaluate every \#iterations & $1$\\
& weight decay & $0.015$ \\
& teacher force ratio & $1.0$ \\
& max iterations & $30$\\
\Xhline{3\arrayrulewidth}
\end{tabular}

\vspace{2mm}

\begin{tabular}{l | c | c}
\Xhline{3\arrayrulewidth}
Model & Parameter & Value \\
\Xhline{0.5\arrayrulewidth}
\multirow{12}{*}{\textsc{Seq2seq}}
& encoder GRU hidden size & $256$ \\
& decoder GRU hidden size & $256$ \\
& attention size & $256$ \\
& word embedding size & $256$ \\
& batchsize & $1$ \\
& optimizer & SGD \\
& initial learning rate & $1e-2$ \\
& \#turns to half learning rate & $5000$ \\
& evaluate every \#iterations & $1000$\\
& weight decay & $0.015$ \\
& teacher force ratio & $0.9$ \\
& max iterations & $185000$\\
\Xhline{3\arrayrulewidth}
\end{tabular}

\vspace{2mm}

\begin{tabular}{l | c | c}
\Xhline{3\arrayrulewidth}
Model & Parameter & Value \\
\Xhline{0.5\arrayrulewidth}
\multirow{13}{*}{\textsc{RetrieveEdit}}
& encoder layers & $2$ \\
& decoder layers & $4$ \\
& hidden size & $256$ \\
& agenda size & $256$ \\
& attention size & $256$ \\
& word embedding size & $300$ \\
& batchsize & $16$ \\
& VAE-kappa & $500$ \\
& ident\_pr & $0.1$ \\
& optimizer & Adam \\
& learning rate & $1e-3$ \\
& max iterations & $1000$\\
& evaluate every \#iterations & 100\\

\Xhline{3\arrayrulewidth}
\end{tabular}
\vspace{-0mm}
\caption{Table of hyperparameters for all models in all experiments respectively. Note that in GPT based models, each iteration means passing through all sentences in the training set, while in GRU+attn and Retrieve-Edit each iteration means passing through a batch in the training set. Also, the vector sizes of all GPT models is equal to the default pretrained GPT2-small model with LM head.\vspace{-2mm}}
\label{hyperparams}
\end{table}

\vspace{-1mm}
\section{Experimental Details}
\label{exp_details}
\vspace{-1mm}

\subsection{Dataset Preprocessing}
\label{preprocessing}

For transfers with additional input to the original sentence (additional information in Info Addition, adjective to emphasize in Adjective Emphasis, etc), we put the additional input at the end of the original sentence separated by a semicolon token. When training Passive To Active and PP Back To Front, due to the low amount of data available, we also include data collected by their reverse operations and swap the source and target. For each transfer, we take all available parallel sentences, and divide them into train, valid and test sets in a $90\%$, $5\%$, $5\%$ ratio. All numerals in the sentences are replaced with a ``NUM'' token when training the baselines.

\begin{table*}[]
\fontsize{9}{11}\selectfont
\setlength\tabcolsep{5.0pt}
\centering
\footnotesize
\begin{tabular}{llccccccc}
\Xhline{3\arrayrulewidth}
Transfer & Baseline Model & \multicolumn{1}{l}{BLEU-1} & \multicolumn{1}{l}{BLEU-2} & \multicolumn{1}{l}{BLEU-3} & \multicolumn{1}{l}{BLEU-4} & \multicolumn{1}{l}{METEOR} & \multicolumn{1}{l}{ROUGE\_L} & \multicolumn{1}{l}{CiDER} \\ \hline
\multirow{3}{*}{To Future Tense} & \textsc{GPT2} & 0.895 & 0.852 & 0.813 & \textbf{0.778} & \textbf{0.540} & 0.899 & 7.709 \\
 & \textsc{Seq2seq} & 0.527 & 0.368 & 0.261 & 0.188 & 0.173 & 0.531 & 1.525 \\
 & \textsc{RetrieveEdit} & \textbf{0.899} & \textbf{0.854} & \textbf{0.815} & \textbf{0.778} & 0.531 & \textbf{0.901} & \textbf{7.731} \\ \cline{2-9}
 & \textsc{Human} & 0.954 & 0.915 & 0.884 & 0.855 & 0.636 & 0.964 & 9.174 \\\hline
\multirow{3}{*}{To Past Tense} & \textsc{GPT2} & 0.836 & 0.776 & 0.722 & 0.674 & 0.484 & 0.842 & 6.700 \\
 & \textsc{Seq2seq} & 0.478 & 0.313 & 0.204 & 0.133 & 0.155 & 0.490 & 1.374 \\
 & \textsc{RetrieveEdit} & \textbf{0.935} & \textbf{0.903} & \textbf{0.873} & \textbf{0.847} & \textbf{0.606} & \textbf{0.933} & \textbf{8.358} \\ \cline{2-9}
 & \textsc{Human} & 0.974 & 0.957 & 0.939 & 0.916 & 0.709 & 0.982 & 9.549 \\ \hline
\multirow{3}{*}{To Present Tense} & \textsc{GPT2} & 0.754 & 0.663 & 0.586 & 0.524 & 0.412 & 0.772 & 5.293 \\
 & \textsc{Seq2seq} & 0.516 & 0.361 & 0.267 & 0.210 & 0.190 & 0.518 & 1.819 \\
 & \textsc{RetrieveEdit} & \textbf{0.909} & \textbf{0.870} & \textbf{0.830} & \textbf{0.793} & \textbf{0.599} & \textbf{0.916} & \textbf{7.987} \\ \cline{2-9}
 & \textsc{Human} & 0.969 & 0.952 & 0.936 & 0.918 & 0.745 & 0.979 & 9.501 \\\hline
 
 \multirow{3}{*}{ADJ or ADV Removal} & \textsc{GPT2} & 0.647 & 0.508 & 0.394 & 0.308 & 0.313 & 0.652 & 3.259 \\
 & \textsc{Seq2seq} & 0.450 & 0.274 & 0.172 & 0.112 & 0.140 & 0.469 & 1.171 \\
 & \textsc{RetrieveEdit} & \textbf{0.897} & \textbf{0.841} & \textbf{0.786} & \textbf{0.731} & \textbf{0.511} & \textbf{0.919} & \textbf{7.461}\\ \cline{2-9}
 & \textsc{Human} & 0.933 & 0.894 & 0.870 & 0.847 & 0.591 & 0.965 & 8.924 \\
\Xhline{3\arrayrulewidth}
\end{tabular}
\caption{Evaluation results on easy transfers.\vspace{-4mm}}
\label{tab:bas_full}
\end{table*}

\begin{table*}[]
\fontsize{9}{11}\selectfont
\setlength\tabcolsep{5.0pt}
\centering
\footnotesize
\begin{tabular}{llccccccc}
\Xhline{3\arrayrulewidth}
Transfer & Baseline Model & \multicolumn{1}{l}{BLEU-1} & \multicolumn{1}{l}{BLEU-2} & \multicolumn{1}{l}{BLEU-3} & \multicolumn{1}{l}{BLEU-4} & \multicolumn{1}{l}{METEOR} & \multicolumn{1}{l}{ROUGE\_L} & \multicolumn{1}{l}{CiDER} \\ \hline
\multirow{3}{*}{PP Front to Back} & \textsc{GPT2} & 0.398 & 0.210 & 0.081 & 0.001 & 0.184 & 0.406 & 0.886 \\
 & \textsc{Seq2seq} & 0.393 & 0.280 & 0.207 & 0.161 & 0.162 & 0.391 & 1.492 \\
 & \textsc{RetrieveEdit} & \textbf{0.541} & \textbf{0.423} & \textbf{0.301} & \textbf{0.176} & \textbf{0.247} & \textbf{0.547} & \textbf{2.536} \\ \cline{2-9}
 & \textsc{Human} & 0.965 & 0.959 & 0.952 & 0.945 & 0.690 & 0.970 & 9.671 \\ \hline
\multirow{3}{*}{PP Back to Front} & \textsc{GPT2} & 0.407 & 0.241 & 0.091 & 0.001 & 0.166 & 0.406 & 0.931 \\
 & \textsc{Seq2seq} & 0.298 & 0.157 & 0.090 & 0.060 & 0.112 & 0.284 & 0.606 \\
 & \textsc{RetrieveEdit} & \textbf{0.649} & \textbf{0.584} & \textbf{0.535} & \textbf{0.491} & \textbf{0.333} & \textbf{0.656} & \textbf{4.667} \\ \cline{2-9}
 & \textsc{Human} & 1.000 & 1.000 & 1.000 & 1.000 & 1.000 & 1.000 & 10.000 \\ \hline
\multirow{3}{*}{PP Removal} & \textsc{GPT2} & 0.763 & 0.700 & 0.645 & 0.593 & 0.419 & 0.787 & 6.012 \\
 & \textsc{Seq2seq} & 0.330 & 0.195 & 0.121 & 0.081 & 0.112 & 0.363 & 1.004 \\
 & \textsc{RetrieveEdit} & \textbf{0.798} & \textbf{0.770} & \textbf{0.739} & \textbf{0.712} & \textbf{0.478} & \textbf{0.846} & \textbf{7.111} \\ \cline{2-9}
 & \textsc{Human} & 0.957 & 0.944 & 0.931 & 0.919 & 0.681 & 0.976 & 9.207 \\\hline
\multirow{3}{*}{Substatement Removal} & \textsc{GPT2} & 0.430 & 0.332 & 0.247 & 0.176 & 0.250 & 0.588 & 3.090 \\
 & \textsc{Seq2seq} & 0.317 & 0.192 & 0.110 & 0.001 & 0.100 & 0.368 & 1.041 \\
 & \textsc{RetrieveEdit} & \textbf{0.706} & \textbf{0.678} & \textbf{0.647} & \textbf{0.607} & \textbf{0.405} & \textbf{0.767} & \textbf{6.183} \\ \cline{2-9} 
 & \textsc{Human} & 0.731 & 0.720 & 0.705 & 0.685 & 0.607 & 0.788 & 7.691 \\ \hline
\multirow{3}{*}{Information Addition} & \textsc{GPT2} & 0.479 & 0.305 & 0.189 & 0.121 & 0.207 & 0.475 & 1.359 \\
 & \textsc{Seq2seq} & 0.345 & 0.180 & 0.094 & 0.053 & 0.098 & 0.335 & 0.632 \\
 & \textsc{RetrieveEdit} & \textbf{0.493} & \textbf{0.396} & \textbf{0.328} & \textbf{0.275} & \textbf{0.284} & \textbf{0.603} & \textbf{3.401} \\ \cline{2-9}
 & \textsc{Human} & 0.846 & 0.762 & 0.690 & 0.624 & 0.521 & 0.892 & 6.863 \\
\Xhline{3\arrayrulewidth}
\end{tabular}
\caption{Evaluation results on medium transfers. \textsc{Info Addition} is especially hard for current models.\vspace{-4mm}}
\label{tab:bas2_full}
\end{table*}

\begin{table*}[]
\fontsize{9}{11}\selectfont
\setlength\tabcolsep{5.0pt}
\centering
\footnotesize
\begin{tabular}{llccccccc}
\Xhline{3\arrayrulewidth}
Transfer & Baseline Model & \multicolumn{1}{l}{BLEU-1} & \multicolumn{1}{l}{BLEU-2} & \multicolumn{1}{l}{BLEU-3} & \multicolumn{1}{l}{BLEU-4} & \multicolumn{1}{l}{METEOR} & \multicolumn{1}{l}{ROUGE\_L} & \multicolumn{1}{l}{CiDER} \\ \hline
\multirow{3}{*}{Active To Passive} & \textsc{GPT2} & 0.476 & 0.329 & 0.238 & 0.189 & 0.216 & 0.464 & 1.820 \\
 & \textsc{Seq2seq} & 0.373 & 0.220 & 0.141 & 0.103 & 0.131 & 0.345 & 0.845 \\
 & \textsc{RetrieveEdit} & \textbf{0.681} & \textbf{0.598} & \textbf{0.503} & \textbf{0.427} & \textbf{0.383} & \textbf{0.663} & \textbf{4.535} \\\cline{2-9}
 & \textsc{Human} &  0.931 & 0.881 & 0.835 & 0.795 & 0.587 & 0.905 & 8.603 \\ \hline
\multirow{3}{*}{Passive To Active} & \textsc{GPT2} & 0.433 & 0.271 & 0.167 & 0.120 & 0.191 & 0.434 & 1.329 \\
 & \textsc{Seq2seq} & 0.339 & 0.214 & 0.160 & 0.132 & 0.126 & 0.331 & 1.062 \\
 & \textsc{RetrieveEdit} & \textbf{0.714} & \textbf{0.659} & \textbf{0.559} & \textbf{0.474} & \textbf{0.397} & \textbf{0.732} & \textbf{5.024} \\ \cline{2-9}
 & \textsc{Human} & 0.977 & 0.962 & 0.942 & 0.919 & 0.685 & 0.973 & 9.409 \\\hline
\multirow{3}{*}{Adjective Emphasis} & \textsc{GPT2} & 0.263 & 0.079 & 0.028 & 0.000 & 0.112 & 0.188 & 0.386 \\
 & \textsc{Seq2seq} & 0.187 & 0.058 & 0.018 & 0.000 & 0.059 & 0.179 & 0.141 \\
 & \textsc{RetrieveEdit} & \textbf{0.387} & \textbf{0.276} & \textbf{0.211} & \textbf{0.164} & \textbf{0.193} & \textbf{0.369} & \textbf{1.679} \\\cline{2-9}
 & \textsc{Human} & 0.834 & 0.753 & 0.679 & 0.611 & 0.522 & 0.811 & 6.796 \\ \hline
\multirow{3}{*}{Verb/Action Emphasis} & \textsc{GPT2} & 0.309 & 0.170 & 0.095 & 0.041 & 0.140 & 0.292 & 0.593 \\
 & \textsc{Seq2seq} & 0.289 & 0.127 & 0.066 & 0.038 & 0.098 & 0.275 & 0.300 \\
 & \textsc{RetrieveEdit} & \textbf{0.416} & \textbf{0.284} & \textbf{0.209} & \textbf{0.148} & \textbf{0.223} & \textbf{0.423} & \textbf{1.778}\\ \cline{2-9}
 & \textsc{Human} & 0.649 & 0.569 & 0.493 & 0.421 & 0.433 & 0.693 & 5.668 \\
\Xhline{3\arrayrulewidth}
\end{tabular}
\caption{Results on hard transfers. Thematic transfers are especially difficult for current models.\vspace{-4mm}}
\label{tab:bas3_full}
\end{table*}

\subsection{Hyperparameters}

The hyperparameters used for all models trained in all experiments is shown in Table~\ref{hyperparams}.

Note that in \textsc{GPT2} based models, each iteration means passing through all sentences in the training set, while in \textsc{Seq2seq} and \textsc{RetrieveEdit} each iteration means passing through a batch in the training set. Also, the vector sizes of all \textsc{GPT2} models is equal to the default pre-trained \textsc{GPT2} (small) model with LM head.

The hyperparameters for \textsc{RetrieveEdit} are the same as the default from the code provided by \citet{hashimoto2018retrieve}\footnote{\url{https://worksheets.codalab.org/worksheets/0x1ad3f387005c492ea913cf0f20c9bb89/}}. The hyperparameters for other models are selected by manual tuning using lowest validation loss.

\subsection{Model Parameters}

Since GPT2 Baselines, \modelshort \ and \textsc{\modelshort-zero}
all uses pretrained GPT2 (small), each of those models have about 124M parameters. Under the hyperparameter settings described above, GRU+attn has about 2.4M parameters. Retrieve-Edit has 51.8M parameters.

\subsection{Training Resources and Time}

All models except \textsc{RetrieveEdit} are run on a single GPU on Google Colab. 
The running time for training \textsc{Seq2seq} for full $185,000$ iterations is about $2$ hours. The training time for \textsc{GPT2} for full $60$ iterations takes between $1$ and $4$ hours (depending on the size of parallel data in the specific transfer), although the best results (in terms of valid loss) can usually be achieved within the first $20$ iterations. The training time for \modelshort\ and \textsc{\modelshort-zero} for full $30$ iterations is about 4 hours on compositional datasets (Tense+Voice, Tense+PP Removal), and the best results can be achieved within the first $10$ iterations. The running time for training each \textsc{RetrieveEdit} model ranges between $40$ minutes and $1$ hour.

\vspace{-1mm}
\section{Full Experimental Results}
\label{experiments_supp}
\vspace{-1mm}

\subsection{Fine-grained Style Transfer}
\label{appx:full_results}

We show complete results of single-style experiments in Table~\ref{tab:bas_full}-~\ref{tab:bas3_full}. We make similar observations that in line with our Hamming distance metric, thematic transfers are especially difficult--all three baselines struggled on this task, which is intuitive because shifting emphasis requires completely different sentence structure changes on different sentences and emphasized words.

Shown below are some examples of thematic transfers done by \textsc{GPT2} and \textsc{RetrieveEdit} model. We found that \textsc{GPT2} and \textsc{Seq2seq} tend to struggle with grammar and word repetitions, while \textsc{RetrieveEdit} sometimes follows the structural edits in the chosen (and often completely unfitting) examples, resulting in malformed outputs (see examples in Appendix~\ref{appx:full_results}). Furthermore, all current methods significantly fall short of human performance especially on hard transfers. Therefore, \datashort\ brings novel challenges that will stimulate future research in modeling fine-grained style changes. Note: in the input, along with the original sentence, the word to emphasize is in {\red{red}}):

\paragraph{Adjective Emphasis}
\begin{mdframed}
\footnotesize
\setlength{\parindent}{0cm}

\textbf{Original Sentence} several other banks have similar applications pending; {\red{similar}}
 
\textbf{Human Annotation:} several other banks have applications pending which are similar

\textbf{\textsc{GPT2}:} other applications applications applications applications applications applications pending

\textbf{\textsc{Seq2seq}:} the bank that the the the the the that was

\textbf{\textsc{RetrieveEdit}:} several applications pending is similar application pending that is

\end{mdframed}

\paragraph{Verb Emphasis}
\begin{mdframed}
\footnotesize
\setlength{\parindent}{0cm}

\textbf{Original Sentence:} i much prefer money i can put my hands on ; {\red{put}}

\textbf{Human Annotation:} putting my hands on money is something i much prefer

\textbf{\textsc{GPT2}:} putting my my my on on on i do do

\textbf{\textsc{Seq2seq}:} the saying that is what we is not to do

\textbf{\textsc{RetrieveEdit}:} the handing of my hands was by something that my hands on it 

\end{mdframed}

\textsc{RetrieveEdit} performed equally well compared to \textsc{GPT2} in some transfers such as To Future Tense and performs significantly better compared to \textsc{GPT2} in most transfers. When qualitatively observing generated sentences, we found that while \textsc{GPT2} can learn syntactic and semantic transfers, they suffer in reconstructing the rest of the sentence (\textit{e.g.} making word repetitions). This was not an issue for \textsc{RetrieveEdit} since it works by editing the sentence from the prototype, not generating the output sentence sequentially. Both \textsc{GPT2} and \textsc{RetrieveEdit} significantly outperform \textsc{Seq2seq} models trained from scratch on all $13$ non-lexical transfers.

\textbf{Human evaluation:} We sampled $10$ transferred sentences from each automatic generations models for each transfer and asked $2$ independent annotators to rate them. We show average results below for one of the hard transfers (Verb Emphasis). From Table~\ref{humanfull}, we found that all approaches fall far short of human performance, which was judged by a separate human as having almost perfect clarity,content, and style metrics. Furthermore, \textsc{GPT2} gets higher style scores while \textsc{RetrieveEdit} excels at grammar and content preservation, which further supports our qualitative observations above.

\begin{table}[t]
\fontsize{9}{11}\selectfont
\setlength\tabcolsep{6.0pt}
\centering
\footnotesize
\begin{tabular}{llccc}
\Xhline{3\arrayrulewidth}
 &  & \multicolumn{1}{l}{Clarity} & \multicolumn{1}{l}{Content} & \multicolumn{1}{l}{Style} \\ \hline
\multirow{4}{*}{Active To Passive} & \textsc{GPT2} & 2.50 & 2.95 & 2.70 \\
 & \textsc{Seq2Seq} & 1.95 & 1.75 & 2.00 \\
 & \textsc{RetrieveEdit} & \textbf{4.05} & \textbf{3.65} & \textbf{4.10} \\ \cline{2-5} 
 & \textsc{Human} & 5.00 & 5.00 & 4.55 \\ \hline
\multirow{4}{*}{To Future} & \textsc{GPT2} & 4.65 & \textbf{4.85} & \textbf{4.80} \\
 & \textsc{Seq2Seq} & 2.05 & 2.20 & 3.25 \\
 & \textsc{RetrieveEdit} & \textbf{4.70} & 4.70 & 4.35 \\ \cline{2-5} 
 & \textsc{Human} & 5.00 & 5.00 & 5.00 \\ \hline
\multirow{4}{*}{ADJ/ADV Removal} & \textsc{GPT2} & 2.65 & 3.50 & \textbf{4.40} \\
 & \textsc{Seq2Seq} & 2.50 & 1.45 & 3.10 \\
 & \textsc{RetrieveEdit} & \textbf{4.65} & \textbf{4.45} & 4.25 \\ \cline{2-5} 
 & \textsc{Human} & 4.95 & 4.95 & 4.25 \\ \hline
\multirow{4}{*}{Substatement Removal} & \textsc{GPT2} & 3.05 & 3.15 & 3.95 \\
 & \textsc{Seq2Seq} & 3.30 & 2.05 & 3.75 \\
 & \textsc{RetrieveEdit} & \textbf{4.30} & \textbf{3.65} & \textbf{4.20} \\ \cline{2-5} 
 & \textsc{Human} & 5.00 & 5.00 & 3.55 \\ \hline
\multirow{4}{*}{Info Add} & \textsc{GPT2} & 2.15 & \textbf{2.55} & \textbf{3.05} \\
 & \textsc{Seq2Seq} & 2.70 & 1.35 & 1.60 \\
 & \textsc{RetrieveEdit} & \textbf{4.00} & \textbf{2.55} & 2.75 \\ \cline{2-5} 
 & \textsc{Human} & 5.00 & 4.80 & 4.70 \\ \hline
\multirow{4}{*}{ADJ emph} & \textsc{GPT2} & 1.30 & 2.65 & 2.85 \\
 & \textsc{Seq2Seq} & 2.65 & 1.05 & 1.00 \\
 & \textsc{RetrieveEdit} & \textbf{4.15} & \textbf{3.05} & \textbf{3.00} \\ \cline{2-5} 
 & \textsc{Human} & 4.55 & 4.75 & 4.75 \\ \hline
\multirow{4}{*}{VB emph} & \textsc{GPT2} & 1.60 & 2.20 & \textbf{4.05} \\
 & \textsc{Seq2Seq} & 3.85 & 1.45 & 1.25 \\
 & \textsc{RetrieveEdit} & \textbf{4.15} & \textbf{2.65} & 2.20 \\ \cline{2-5} 
 & \textsc{Human} & 4.70 & 4.45 & 5.00 \\ 
 
 \Xhline{3\arrayrulewidth}
\end{tabular}
\caption{Human evaluation for single atomic style transfer on 7 selected transfers (the 7 transfers with BLEU scores appearing in main part of paper). The result shows that on harder transfers, all approaches fall short of human performance, and that \textsc{GPT2} excels at style while \textsc{RetrieveEdit} is better at grammar and content preservation.\vspace{-2mm}}
\label{humanfull}
\end{table}

\subsection{Compositional Style Transfer}
\label{comp_examples}

\begin{table*}[]
\fontsize{9}{11}\selectfont
\begin{tabular}{l|l|l}
\Xhline{3\arrayrulewidth}
Transfer & To Future + Passive To Active & To Past + PP Removal \\ \hline
Source Sentence & \begin{tabular}[c]{@{}l@{}}NUM \% was risen by sales to NUM billion from NUM billion\end{tabular} & \begin{tabular}[c]{@{}l@{}}the bond market was unmoved\\ by the economic statistics\end{tabular} \\ \hline
Target Sentence & sales will rise NUM \% to NUM billion from NUM billion & the bond market is unmoved \\ \hline
\textsc{SeqGPT} & willalesalesalesales to billion from from NUM billion & the bond market is is \\ \hline
\textsc{\modelshort-zero} & \begin{tabular}[c]{@{}l@{}}NUM \% \% \% risen risen sales \\ sales NUM NUM from NUM billion\end{tabular} & \begin{tabular}[c]{@{}l@{}}the bond market is unmoved\\ by the economic statistics\end{tabular} \\
\hline
\modelshort & {\color{gg}sales will rise NUM \% to NUM billion from NUM billion} & {\color{gg}the bond market is unmoved} \\
\Xhline{3\arrayrulewidth}
\end{tabular}
\caption{2 examples of composition transfers generated by \modelshort, \textsc{SeqGPT} and \textsc{\modelshort-zero}. \modelshort \ successfully models compositional transfers across multiple styles.\vspace{-4mm}}
\label{tab:comp_example_supp}
\end{table*}

We present full results on compositional style transfer in Table~\ref{tab:comp_full} and show more examples of compositional transfers done by \modelshort, \textsc{\modelshort-zero}, and \textsc{SeqGPT} in Table~\ref{tab:comp_example_supp}. \modelshort\ significantly outperforms existing methods in all compositional style transfer tasks in both datasets. This is expected, as \modelshort\ is trained on the full compositional datasets, while \textsc{\modelshort-zero} is only trained on part of the compositional dataset and each part of \textsc{SeqGPT} is trained on single-transfer parallel data. Qualitatively, we observed that \modelshort\ is able to perform each required transfer at the same time, producing outputs with relatively low reconstruction error compared to the other two methods.

\begin{table*}[]
\fontsize{9}{11}\selectfont
\setlength\tabcolsep{2.0pt}
\centering
\footnotesize
\begin{tabular}{lllccccccc}
\Xhline{3\arrayrulewidth}
Dataset & Transfers & Model & BLEU-1 & BLEU-2 & BLEU-3 & BLEU-4 & METEOR & ROUGE\_L & CiDER \\ \hline
\multirow{18}{*}{\begin{tabular}[c]{@{}l@{}}Tense\\ +\\ Voice\end{tabular}} & \multirow{3}{*}{\begin{tabular}[c]{@{}l@{}}ToPast+\\ ActiveToPassive\end{tabular}} & \textsc{SeqGPT} & 0.332 & 0.155 & 0.057 & 0.024 & 0.144 & 0.300 & 0.636 \\
 & & \textsc{\modelshort-zero} & 0.337 & 0.163 & 0.075 & 0.029 & 0.154 & 0.283 & 0.760 \\
 & & \modelshort & \textbf{0.409} & \textbf{0.238} & \textbf{0.133} & \textbf{0.064} & \textbf{0.180} & \textbf{0.378} & \textbf{1.029} \\
 \cline{2-10}
 & \multirow{3}{*}{\begin{tabular}[c]{@{}l@{}}ToFuture+\\ ActiveToPassive\end{tabular}} & \textsc{SeqGPT} & 0.391 & 0.222 & 0.120 & 0.065 & 0.167 & 0.373 & 0.866 \\
 & & \textsc{\modelshort-zero} & 0.419 & 0.243 & 0.114 & 0.047 & 0.209 & 0.325 & 1.238 \\
 & & \modelshort & \textbf{0.496} & \textbf{0.340} & \textbf{0.240} & \textbf{0.185} & \textbf{0.217} & \textbf{0.479} & \textbf{1.800} \\
 \cline{2-10}
 & \multirow{3}{*}{\begin{tabular}[c]{@{}l@{}}ToFuture+\\ PassiveToActive\end{tabular}} & \textsc{SeqGPT} & 0.401 & 0.212 & 0.097 & 0.048 & 0.163 & 0.385 & 0.888 \\
 & & \textsc{\modelshort-zero} & 0.399 & 0.245 & 0.123 & 0.047 & 0.212 & 0.349 & 1.075 \\ 
 & & \modelshort & \textbf{0.528} & \textbf{0.364} & \textbf{0.259} & \textbf{0.197} & \textbf{0.234} & \textbf{0.524} & \textbf{2.020} \\
 \cline{2-10}
 & \multirow{3}{*}{\begin{tabular}[c]{@{}l@{}}ToPast+\\ PassiveToActive\end{tabular}} & \textsc{SeqGPT} & 0.381 & 0.210 & 0.098 & 0.045 & 0.156 & 0.368 & 0.876 \\
 & & \textsc{\modelshort-zero} & 0.365 & 0.181 & 0.073 & 0.025 & 0.156 & 0.343 & 0.752 \\ 
 & & \modelshort & \textbf{0.474} & \textbf{0.297} & \textbf{0.175} & \textbf{0.099} & \textbf{0.206} & \textbf{0.473} & \textbf{1.513} \\
 \cline{2-10} 
 & \multirow{3}{*}{\begin{tabular}[c]{@{}l@{}}ToPresent+\\ PassiveToActive\end{tabular}} & \textsc{SeqGPT} & 0.348 & 0.189 & 0.085 & 0.037 & 0.142 & 0.343 & 0.745 \\
 & & \textsc{\modelshort-zero} & 0.424 & 0.257 & 0.118 & 0.046 & 0.208 & 0.389 & 1.025 \\
 & & \modelshort & \textbf{0.523} & \textbf{0.366} & \textbf{0.264} & \textbf{0.210} & \textbf{0.243} & \textbf{0.522} & \textbf{2.118} \\
 \cline{2-10} 
 & \multirow{3}{*}{\begin{tabular}[c]{@{}l@{}}ToPresent+\\ ActiveToPassive\end{tabular}} & \textsc{SeqGPT} & 0.396 & 0.256 & 0.177 & 0.136 & 0.179 & 0.384 & 1.209 \\
 & & \textsc{\modelshort-zero} & 0.445 & 0.254 & 0.120 & 0.059 & 0.212 & 0.348 & 1.271 \\
 & & \modelshort & \textbf{0.503} & \textbf{0.358} & \textbf{0.271} & \textbf{0.223} & \textbf{0.233} & \textbf{0.491} & \textbf{2.118} \\
 \hline
 \multirow{9}{*}{\begin{tabular}[c]{@{}l@{}}Tense\\ +\\ PP \\ Removal\end{tabular}} & \multirow{3}{*}{\begin{tabular}[c]{@{}l@{}}ToFuture+\\ PPRemoval\end{tabular}} & \textsc{SeqGPT} & 0.722 & 0.644 & \textbf{0.581} & \textbf{0.524} & 0.385 & \textbf{0.755} & 5.562 \\
 & & \textsc{\modelshort-zero} & 0.465 & 0.335 & 0.221 & 0.137 & 0.313 & 0.496 & 1.907 \\
 & & \modelshort & \textbf{0.738} & \textbf{0.652} & 0.578 & 0.518 & \textbf{0.393} & 0.755 & \textbf{5.289} \\
 \cline{2-10} 
 & \multirow{3}{*}{\begin{tabular}[c]{@{}l@{}}ToPast+\\ PPRemoval\end{tabular}} & \textsc{SeqGPT} & 0.714 & 0.640 & 0.573 & 0.510 & 0.374 & 0.724 & 5.152 \\
 & & \textsc{\modelshort-zero} & 0.542 & 0.389 & 0.268 & 0.182 & 0.314 & 0.535 & 2.103 \\
 & & \modelshort & \textbf{0.772} & \textbf{0.695} & \textbf{0.624} & \textbf{0.564} & \textbf{0.421} & \textbf{0.775} & \textbf{5.585} \\
 \cline{2-10} 
 & \multirow{3}{*}{\begin{tabular}[c]{@{}l@{}}ToPresent+\\ PPRemoval\end{tabular}} & \textsc{SeqGPT} & 0.618 & 0.518 & 0.435 & 0.368 & 0.338 & 0.663 & 4.119 \\
 & & \textsc{\modelshort-zero} & 0.545 & 0.393 & 0.269 & 0.184 & 0.323 & 0.539 & 2.017 \\
 & & \modelshort & \textbf{0.709} & \textbf{0.609} & \textbf{0.523} & \textbf{0.446} & \textbf{0.718} & \textbf{0.718} & \textbf{4.588} \\
\Xhline{3\arrayrulewidth}
\end{tabular}
\caption{Results on compositions of transfers using sequentially applying \textsc{GPT2} (\textsc{SeqGPT}), \textsc{\modelshort-zero} (adding compositional model but not compositional data) and \modelshort\ (with both compositional model and data). The result shows that \modelshort\ significantly outperforms the other two methods, and zero-shot remains challenging as \textsc{\modelshort-zero} does not perform very well in comparison. \vspace{-4mm}}
\label{tab:comp_full}
\end{table*}

We also present full comparisons of \modelshort\ and \textsc{GPT2} on single style transfer are in Table~\ref{tab:single_full}. We observe that \modelshort\ can often perform single transfers better than \textsc{GPT2} trained specifically for that one task, while in the rest of the cases the \modelshort\ and \textsc{GPT2} has nearly the same performance. Therefore, \modelshort\ has leveraged compositional structure and data to perform strongly on multiple single and compositional transfers with just one model.

\begin{table*}[]
\fontsize{9}{11}\selectfont
\setlength\tabcolsep{5.0pt}
\centering
\footnotesize
\begin{tabular}{llccccccc}
\Xhline{3\arrayrulewidth}
Transfer & Model & BLEU-1 & BLEU-2 & BLEU-3 & BLEU-4 & METEOR & ROUGE\_L & CiDER \\ \hline
To Future Tense & \textsc{GPT2} & \textbf{0.895} & \textbf{0.851} & \textbf{0.812} & \textbf{0.777} & \textbf{0.539} & \textbf{0.898} & \textbf{7.708} \\
 & \modelshort\ (TV) & 0.727 & 0.614 & 0.614 & 0.450 & 0.362 & 0.731 & 4.386 \\
 & \modelshort\ (TP) & 0.810 & 0.731 & 0.663 & 0.606 & 0.446 & 0.818 & 6.026 \\
 \hline
\multirow{3}{*}{To Past Tense} & \textsc{GPT2} & \textbf{0.835} & \textbf{0.776} & \textbf{0.721} & \textbf{0.673} & 0.484 & \textbf{0.842} & 6.699 \\ 
 & \modelshort\ (TV) & 0.694 & 0.586 & 0.494 & 0.420 & 0.353 & 0.700 & 4.051 \\
 & \modelshort\ (TP) & 0.834 & 0.771 & 0.718 & 0.672 & \textbf{0.486} & 0.841 & \textbf{6.704} \\
 \hline
\multirow{3}{*}{To Present Tense} & \textsc{GPT2} & 0.753 & 0.662 & 0.586 & 0.523 & 0.412 & 0.772 & 5.293 \\
 & \modelshort\ (TV) & 0.733 & 0.635 & 0.553 & 0.488 & 0.387 & 0.744 & 4.742 \\
 & \modelshort\ (TP) & \textbf{0.826} & \textbf{0.755} & \textbf{0.691} & \textbf{0.637} & \textbf{0.491} & \textbf{0.831} & \textbf{6.315} \\
 \hline
\multirow{2}{*}{ActiveToPassive} & \textsc{GPT2} & \textbf{0.475} & \textbf{0.329} & \textbf{0.238} & \textbf{0.189} & 0.216 & \textbf{0.463} & \textbf{1.820} \\
 & \modelshort\ (TV) & 0.472 & 0.324 & 0.232 & 0.179 & \textbf{0.216} & 0.454 & 1.790 \\
 \hline
\multirow{2}{*}{PassiveToActive} & \textsc{GPT2} & 0.433 & 0.271 & 0.167 & 0.120 & 0.191 & 0.434 & 1.329 \\
 & \modelshort\ (TV) & \textbf{0.506} & \textbf{0.345} & \textbf{0.243} & \textbf{0.184} & \textbf{0.229} & \textbf{0.505} & \textbf{1.958} \\
 \hline
\multirow{2}{*}{PP Removal} & \textsc{GPT2} & \textbf{0.763} & \textbf{0.700} & \textbf{0.645} & \textbf{0.593} & 0.419 & \textbf{0.786} & \textbf{6.011} \\
 & \modelshort\ (TP) & 0.760 & 0.698 & 0.639 & 0.585 & \textbf{0.420} & 0.772 & 5.783 \\
\Xhline{3\arrayrulewidth}
\end{tabular}
\caption{Comparing single transfer performances between \modelshort \ and \textsc{GPT2} baselines (where TV indicates the \modelshort\ is trained on Tense+Voice dataset and TP indicates the \modelshort\ is trained on Tense+PP Removal dataset). The result shows that \modelshort\ can perform multiple single style transfers with similar performance to \textsc{GPT2} trained specifically for that one transfer, and sometimes even outperforms \textsc{GPT2}.\vspace{-4mm}}
\label{tab:single_full}
\end{table*}

\vspace{-1mm}
\section{Mitigating Social Biases: Qualitative Evaluation}
\label{bias_supp}
\vspace{-1mm}

We created two prompts ``The man worked as'' and ``The woman worked as'', and generated $50$ sentences with each prompt from \textsc{GPT2}. Next, we determine biased words by taking the $1,000$ closest vectors in GloVe word embeddings~\cite{pennington2014glove} to ``man'' and ``woman''.  Then, we determine a sentence as biased if the phrase describing the occupation in the sentences contains any biased words. With this standard, we found that $21$ out of $50$ sentences for man and $28$ out of $50$ sentences are biased. Then, we replaced the occupations in these $49$ biased sentences with occupations sampled uniformly randomly from all $100$ generated sentences, and then asked two independent human annotators to evaluate the $49$ replaced sentences on a five-point scale of \textbf{Significantly More Biased, Slightly More Biased, The Same, Slightly Less Biased}, and \textbf{Significantly Less Biased}. On average, the annotators reported $22$ sentences being significantly less biased compared to before the replacements, while all other sentences are either slightly less biased or neutral. The full results of this experiment are shown in Table~\ref{biasexperiment}. A few examples that were deemed \textbf{Significantly Less Biased} by both annotators are shown in Table~\ref{last_bias_example}.

\begin{table}[]
\fontsize{9}{11}\selectfont
\centering

\setlength\tabcolsep{3.5pt}

\begin{tabular}{l |cc|c}
\Xhline{3\arrayrulewidth}
& Male context & Female context & Total \\ \hline
Biased & $21$ & $28$ & $49$ \\
Not Biased & $29$ & $22$ & $51$  \\ \hline
Total & $50$ & $50$ & $100$ \\
\Xhline{3\arrayrulewidth}
\end{tabular}

\vspace{2mm}

\begin{tabular}{l | c}
\Xhline{3\arrayrulewidth}
Category & Number \\
\Xhline{0.5\arrayrulewidth}
Significantly more biased & $0$ \\
Slightly more biased & $0$ \\
Little or no change in bias & $22$ \\
Slightly less biased & $5$ \\
Significantly less biased & $\mathbf{22}$ \\ \hline
Total & $49$ \\
\Xhline{3\arrayrulewidth}
\end{tabular}

\vspace{-0mm}
\caption{\textbf{Top table:} Human annotators found that $21$ out of $50$ sentences generated by \textsc{GPT2} on ``The man worked as'' and $28$ out of $50$ sentences generated by \textsc{GPT2} on ``The woman worked as'' to exhibit gender bias. \textbf{Bottom table:} Out of the $49$ biased sentences, after using style transfer to replace occupations with randomly sampled ones, human annotators found that $22$ of them are significantly less biased while the rest are either slightly less biased or neutral. \vspace{-2mm}}
\label{biasexperiment}
\end{table}

\begin{table}[]
\fontsize{9}{11}\selectfont
\setlength\tabcolsep{5.0pt}
\centering
\footnotesize
\begin{tabular}{ll}
\Xhline{3\arrayrulewidth}

% \begin{table}[]
% \begin{tabular}{ll}
Text Generated by \textsc{GPT2} & Text after occupation replacement \\ \hline
\begin{tabular}[c]{@{}l@{}}The man worked as a \textbf{security guard}\\ for the family's hotel until the weekend.\end{tabular} & \begin{tabular}[c]{@{}l@{}}The man worked as a \textbf{receptionist}\\ for the family's hotel until the weekend.\end{tabular} \\
The man worked as a driver for his father. & The man worked as a \textbf{personal care aide} for his father. \\
\begin{tabular}[c]{@{}l@{}}The woman worked as a \textbf{maid} at a\\ resort in a small town.\end{tabular} & \begin{tabular}[c]{@{}l@{}}The woman worked as a \textbf{driver} at a resort\\ in a small town.\end{tabular} \\
The woman worked as a \textbf{nurse} at a facility. & The woman worked as a \textbf{construction worker} at a facility. \\
\Xhline{3\arrayrulewidth}
\end{tabular}
\caption{Examples of sentences generated by \textsc{GPT2} and with occupation replacements that are rated as ``Significantly Less Biased'' after the change by human annotators. \vspace{-4mm}}
\label{last_bias_example}
\end{table}

\end{document}